\documentclass{article}

 \usepackage[main, final]{neurips_2026}

\usepackage[utf8]{inputenc} 
\usepackage[T1]{fontenc}    
\usepackage{hyperref}       
\usepackage{url}            
\usepackage{booktabs}       
\usepackage{amsfonts}       
\usepackage{nicefrac}       
\usepackage{microtype}      
\usepackage{xcolor}         

\newcommand{\promptText}{Sports Academy has found “The Mamba Mentality.” The 96,000-square-foot training facility in Thousand Oaks announced a partnership with Kobe Bryant.
The 96,000-square-foot holistic athletic training facility in Thousand Oaks announced a partnership with former Lakers superstar Kobe Bryant on Monday morning. }

\newcommand{\mamarkdeltatwo}{
film for “Dear Basketball.” He is also an executive producer on the “What Doesn’t Kill You” Netflix series.
Sports Academy also operates training facilities in Beverly Hills, Santa Barbara and Newport Beach. A fifth center is set to open in Westlake Village in early February 2020.
Sports Leagues and Activities
Conejo Hills Elementary School in Thousand...
Conejo High School District adopts new...
Thousand Oaks school district reinstates mask mandate
Calabasas school district to reinstate mask mandate
Firefighters battle blaze in Simi Valley; evacuations ordered in Thousand Palms
Thousand Palms fire update: 5,563 acres, 10
Thousand palms wildfire: 5,663 acres, no structures lost, evacuations lifted
Thousand Hills Fire: Cause of wildfire under investigation; mandatory evacuations lifted for most residents
Thousand-Hills Fire: Cause under investigation; mandated evacuations lifted but evacuations
}

\newcommand{\mpacfour}{short.
This article is written by Mike Coppinger from Los Angeles Daily News and was legally licensed via the Tribune Content Agency through the NewsCred publisher network. Please direct all licensing questions to [email protected] .
Next Post: The top 265 hotel openings, renovations set to launch in 2023
This article is really interesting. Thanks for the great post!
Governors Island is opening a sports arena with artificial turf and more.
Based in a city that boasts an impressive number of parks and recreation centers, Washington DC still...
Tennis courts are among the popular sports amenities in resorts.
Tennis courts have long been staple sports amenities in vacation homes, luxury hotels, and resorts. Today’s...
Leading Resort Operators Launch New Brand Campaign, ‘Live the High Life’
Leading resort operator Great Western Resorts has united with its parent, Highgate Hotels, as well as...
Tennis court construction is easy if you are properly prepared.
Tennis court surfaces come in a variety of different}

\newcommand{\plusfour}{video.
This article is written by Samantha Mascarenas from The Ventura County Star and was legally licensed via the Tribune Content Agency through the NewsCred publisher network. Please direct all licensing questions to [email protected].
Copyright 2019, Ventura County Star. All rights reserved. From The Los Angeles Daily News, California. Distributed by Tribune Content Agency, LLC.
Want to be the first to hear what's new in the fitness space and in our industry? Subscribe to the Sweat Equity Podcast!
©Copyright 2107. All rights Reserved. FitLife Brands, Inc.
"The Sweat Equality Podcast" and "Sweat Equality" is the property and trademark of FitLife Brends,Inc. All rights Resereved.
You can't do it alone, let us help you. We can do the heavy lifting for you, so you'll have more time to focus on your business.
Tell us what you need and we'll take care of it.
You're not alone.}

\newcommand{\rsbhfour}{He'll host a workout and clinic at the academy this spring and will be involved at the academy through social media promotions.
He's already been involved through social media promos.
“As a professional athlete who has spent nearly two decades training at world-class facilities, I know the value of sports academies and the importance they have in the lives of youth,” Bryant said in a statement.
“Sports Academy is a state of the art, all-inclusive facility that offers unparalleled opportunities to develop the body and the mind.
“My team and I are thrilled to have our brand aligned alongside the most progressive training facility in the world and the entire MAMBA team," said Faulkner in a statement. "...We look forward to this partnership, in which we can share the same values and principles with each other."
Want to see more from our team at the Ventura County Star?
You must be a digital subscriber to view this content.
You must be registered with your Ventura County Star account to view this content!
You must be signed in to your Ventura County account to}

\usepackage{amsmath}
\usepackage{amssymb}
\usepackage{mathtools}
\usepackage{amsthm}

\usepackage{booktabs}
\usepackage{multirow, multicol}
\usepackage{pifont}
\newcommand{\cmark}{\ding{51}}%
\newcommand{\xmark}{\ding{55}}%
\usepackage{threeparttable} 
\usepackage{makecell}
\usepackage{xcolor}
\usepackage{colortbl}
\usepackage{graphicx}
\usepackage{wrapfig}
\usepackage{subcaption}
\usepackage{enumitem}

\usepackage[ruled]{algorithm2e}

\usepackage{amsmath, amsthm}
\usepackage{amssymb}
\usepackage{mathtools}
\newtheorem{theorem}{Theorem}

\newtheorem{definition}{Definition}
\newtheorem{assumption}{Assumption}
\newtheorem{proposition}{Proposition}
\newtheorem{remark}{Remark}

\newcommand{\eqautoref}[1]{\hyperref[#1]{Equation (\ref{#1})}}
\newcommand{\ineqautoref}[1]{\hyperref[#1]{Inequality (\ref{#1})}}

\newcommand{\algautoref}[1]{\hyperref[#1]{Algorithm~\ref{#1}}}
\newcommand{\apdxautoref}[1]{\hyperref[#1]{Appendix~\ref{#1}}}

\newcommand{\sectionautoref}[1]{\hyperref[#1]{Section~\ref{#1}}}

\newcommand{\formulaautoref}[1]{\hyperref[#1]{Formula~\ref{#1}}}
\newcommand{\problemautoref}[1]{\hyperref[#1]{Problem~\ref{#1}}}

\newcommand{\lineautoref}[1]{\hyperref[#1]{line~\ref{#1}}}

\title{Majority Bit-Aware Watermarking for Large Language Models}

\author{%
  Jiahao Xu$^{1, 2}$\thanks{This work was done during Jiahao's internship at the Oak~Ridge~National~Laboratory.} \quad Rui Hu$^1$ \quad Zikai Zhang$^1$ \quad Olivera Kotevska$^2$ \\
  $^1$University of Nevada, Reno\\
  $^2$Oak Ridge National Laboratory\\
  \texttt{\{jiahaox, ruihu, zikaiz\}@unr.edu} \quad \texttt{kotevskao@ornl.gov} \\
}

\begin{document}

\maketitle

\begin{abstract}
  The growing deployment of Large Language Models (LLMs) has raised concerns about their misuse in generating harmful or deceptive content. To address this issue, watermarking methods have been proposed to embed identifiable multi-bit messages into generated text for misuse tracing. However, existing methods often suffer from a fundamental trade-off between text quality and decoding accuracy. In particular, they have to restrict the size of the preferred token set (i.e., green list) during encoding to maintain a detectable watermark signal for decoding, which inevitably degrades generation quality. To improve this trade-off, we propose a novel message encoding paradigm called \textit{majority bit-aware encoding}, which relaxes the watermark signal strength from the green list size. This strategy allows for a strong watermark signal to be preserved in generated texts even when using a large green list. We introduce two instantiations of this paradigm: MajorMark and MajorMark$^{+}$, where the latter is specifically optimized for long messages. Extensive experiments on state-of-the-art LLMs demonstrate that our methods achieve higher decoding accuracy and superior text quality compared to prior baselines. The code is at \url{https://github.com/JiiahaoXU/MajorMark}.
\end{abstract}

\section{Introduction}

The widespread deployment of Large Language Models (LLMs) has fundamentally reshaped numerous domains, from education and customer support to sophisticated content generation, owing to their capacity for highly fluent and contextually relevant text. 
However, this powerful capability simultaneously introduces considerable security and ethical challenges. Malicious users can leverage LLMs to produce harmful content, such as fake news, phishing emails, and fraudulent reviews~\citep{rsbh, zhang2025character}. To mitigate these risks, developing watermarking techniques is essential. By embedding an imperceptible yet verifiable signal within generated text, watermarking enables effective origin tracing and attribution of misuse~\citep{liu2024survey, wu2025survey, pan2024markllm}.
A typical text watermarking method comprises an \textit{encoder} that embeds a specific message into the generated content and a corresponding \textit{decoder} designed to extract this message. 
This work focuses on a common watermarking scenario~\citep{mpac, rsbh, stealthink}, where a cloud-based LLM-as-a-Service provider adopts \textbf{\textit{multi-bit watermarking}} to uniquely mark each output for a user prompt.
%
%
The decoder can then recover the embedded message from unverified text. 

The first LLM watermarking scheme, KGW~\citep{kgw}, was designed to encode a \textit{zero-bit message}--a yes/no signal indicating that an LLM generated the text. 
During encoding, at each token generation step, KGW pseudo-randomly selects a \textit{green list} $\mathcal{G} \subseteq \mathcal{V}$ from the vocabulary and strategically boosts the logits of the tokens in $\mathcal{G}$ by a watermarking bias $\delta$. This increases their probability of being sampled as the next token. A larger $\delta$ produces a stronger watermark but often results in noticeable degradation in the quality of the generated text. 
During decoding, KGW reconstructs the green list for each token within the target text and computes the proportion of tokens that fall into their respective green lists. If this proportion is statistically significant (measured via a $z$-test), the target text is verified as watermarked and, consequently, generated by an LLM. 
%

However, \textit{zero-bit watermarking} methods are inherently limited in our setting, since service providers may need to embed more complex information, such as user identifiers, model versions, or other metadata. To address this limitation, recent studies have explored \textbf{\textit{multi-bit watermarking}}, which is designed to embed a binary message $\mathbf{m}$ of length $b \geq 2$. Existing methods can be broadly divided into two categories: perturbation-based methods~\citep{mpac, depthw, ctwl, cycleshift, rsbh} and distortion-free methods~\citep{kuditipudi2024robust, stealthink, feng2025bimark}. In this work, we primarily focus on perturbation-based methods, as they typically achieve higher decoding accuracy. However, such improvements often come at the cost of reduced quality in the generated watermarked text. Specifically, two key watermarking parameters affect the quality of watermarked text in perturbation-based methods: the green list ratio, defined as $\gamma = |\mathcal{G}|/|\mathcal{V}|$, and the watermarking bias $\delta$ used to boost the logits of green-list tokens. In this work, we study the impact of the green list ratio $\gamma$ while fixing the watermarking bias. Intuitively, a smaller $\gamma$ can degrade the quality of watermarked text, as a narrower green list provides fewer contextually suitable token choices during generation. In contrast, under a fixed watermarking bias, a smaller $\gamma$ can improve decoding accuracy, since the decoding process in existing methods relies on \textit{counting the frequency of tokens} that appear in the corresponding green lists. Therefore, existing methods exhibit an empirical trade-off between \textit{watermarked text quality and message decoding accuracy}, as shown in Section~\ref{sec: problem_and_motivation}.

To improve this trade-off, we propose a novel \textbf{\textit{majority bit-aware encoding strategy}} that relaxes the watermark signal strength from the size of the green list, allowing a large green list to still carry a strong watermark signal. Specifically, we partition the vocabulary into $b$ disjoint shards, where each shard corresponds to a specific bit in the message. During encoding, the green list is constructed by uniting all shards that correspond to the message's majority bit. This strategy inherently guarantees a green list ratio of $\gamma \geq 0.5$, which is critical for preserving text quality. Crucially, the message can now be represented via the specific selection of green shards. Consequently, the decoding objective shifts from analyzing token frequency to inferring which specific shards were boosted during encoding.

Based on the proposed majority bit-aware encoding strategy, we develop MajorMark, a \underline{\textit{Major}}ity bit-aware water\underline{\textit{Mark}}ing method. MajorMark directly leverages majority bit-aware encoding, yielding an expected green list ratio $\mathbb{E}_\mathbf{m}[\gamma] \approx 0.5 + {1}/{\sqrt{2 \pi b}}$. It is equipped with a clustering-based decoding strategy that operates on shard-level occurrence statistics, enabling fine-grained message decoding. Furthermore, we propose MajorMark$^{+}$ to better preserve text quality for long messages. MajorMark$^{+}$ partitions the message into $r$ blocks and applies the majority bit-aware encoding to each block independently, yielding a larger expected green list ratio, $\mathbb{E}_\mathbf{m}[\gamma] \approx 0.5 + {1}/{\sqrt{2 \pi (b/r)}}$. During decoding, MajorMark$^{+}$ deterministically reconstructs each message block by inferring the majority bit and the number of green shards.
Our key contributions are summarized as follows:

\begin{itemize}[leftmargin=*]
    \item We propose a novel \textit{majority bit-aware encoding strategy} for multi-bit watermarking of LLM-generated content. By encoding the message into the selection of green shards, this paradigm effectively relaxes watermark detectability from the green list size, thereby improving the trade-off between text quality and decoding accuracy.

    \item We propose two multi-bit watermarking methods, termed MajorMark and MajorMark$^+$, both of which adopt the majority bit-aware encoding strategy to preserve high text quality. We further design corresponding decoders for each method to ensure accurate message decoding.

    \item We conduct extensive experiments with state-of-the-art open-source LLMs on different downstream tasks. The results show that our methods consistently achieve higher decoding accuracy and text quality than existing multi-bit watermarking approaches.


\end{itemize}

\section{Preliminaries and Existing Works}


\textbf{Large Language Models.} 
An LLM $f$ is an autoregressive model that performs the next-token prediction task. Specifically, given an input prompt $\mathbf{x}_p$, the model generates a sequence of tokens over a predefined vocabulary $\mathcal{V}$, which contains all permissible tokens for constructing natural language responses. At the $t$-th ($t = 1, 2, \dots, T$) generation step, the model takes as input the prompt $\mathbf{x}_p$ and the previously generated tokens $\mathbf{x}_{:t-1} = \{x_1, x_2, \ldots, x_{t-1} \}$, and outputs a logits vector $\ell^{t} = (\dots, f(v \mid \mathbf{x}_p, \mathbf{x}_{:t-1}), \dots)$, where $f(v \mid \mathbf{x}_p, \mathbf{x}_{:t-1})$ represents the logit for a specific token $v \in \mathcal{V}$. These logits are then passed through a $\mathtt{softmax}(\cdot)$ to produce a probability distribution over $\mathcal{V}$, from which the current token $x_t$ is sampled. This process is repeated autoregressively until a termination condition (e.g., maximum length) is met, yielding an output sequence $\mathbf{x}_g$ of length $T$.

\textbf{Zero-Bit Watermarking in LLMs.} The first zero-bit watermarking method for LLMs, KGW~\citep{kgw}, embeds a detectable signal into generated text without encoding an explicit message. At each generation step $t$, the LLM produces a logit vector, after which a pseudo-random seed is derived by hashing a secret key $k$ with the previous token $x_{t-1}$. This seed deterministically permutes and partitions the vocabulary $\mathcal{V}$ into a green list $\mathcal{G}$ and a red list $\mathcal{R}$. A positive watermarking bias $\delta$ is then added to the logits of tokens in $\mathcal{G}$, increasing their sampling probability after the softmax operation. While a larger $\delta$ strengthens the watermark signal, it also more severely perturbs the token distribution, which can degrade text quality. This procedure is applied at every token generation step, producing a watermarked sequence. To detect a watermark in an unverified text, the decoder reconstructs the green list for each token and performs a $z$-test on the frequency of tokens falling into their corresponding green lists. A statistically significant excess over the expected frequency indicates the presence of a watermark. Subsequent works have extended this framework to improve decoding accuracy and text quality~\citep{kirchenbauer2024on, kuditipudi2024robust, wang2025morphmark, postmark, giboulot2024watermax, liu2024a, piet2025markmywords, christ2024undetectable, munyer2024deeptextmark}, as well as to enhance robustness against post-generation text editing~\citep{liu2024an, liu2024a, hou2024semstamp, dabiriaghdam2025simmark}.

\setlength{\intextsep}{0mm}%
\begin{wraptable}{r}{0.48\textwidth} 
  \centering
  \caption{Comparison of representative multi-bit LLM watermarking methods. 
  $\mathbb{E}_\mathbf{m}[\gamma]$ denotes the expected green list ratio over all possible messages.
  $\times |\mathbf{x}|$ indicates that the decoding process requires how many enumerations over all tokens in the input text $\mathbf{x}$.
  ``Eff.?'' reflects whether the decoding procedure is computationally efficient.}
  \label{tab:summarize_method}
  \scalebox{0.74}{
  \begin{threeparttable}
    \begin{tabular}{l|c|cc}
    \toprule
    \multirow{2}[2]{*}{\textbf{Method}} & \multicolumn{1}{c|}{\textbf{Encoding}} & \multicolumn{2}{c}{\textbf{Decoding}}  \\
          & $\mathbb{E}_\mathbf{m}[\gamma]$    & $\times |\mathbf{x}|$ & Eff.?  \\
    \midrule
    CTWL~\cite{ctwl}$^\dagger$ & $-$  & $2^b$ & \xmark\\
    CycleShift~\cite{cycleshift} & $=0.25$  & $2^b$ & \xmark \\
    DepthW~\cite{depthw} & $=0.50$  & $2^b$ & \xmark \\
    RSBH~\cite{rsbh}$^\ddagger$  &   $=0.50$    &  $2^{d}$ & \cmark \\
    MPAC~\cite{mpac}  &   $=0.25$     & $1$ & \cmark  \\
    \midrule
    \rowcolor[rgb]{.9,.9,.9}MajorMark  &   $\approx0.50 + 1/\sqrt{2 \pi b}$     & $2$ & \cmark  \\
    \rowcolor[rgb]{.9,.9,.9}MajorMark$^+$$^\sharp$  &   $\approx0.50 + 1/\sqrt{2 \pi (b/r)}$     & $b/r-1$ & \cmark  \\
    \bottomrule
    \end{tabular}
    \begin{tablenotes}
      \footnotesize
      \item[$\dagger$] CTWL determines $\gamma$ dynamically by a proxy LLM.
      \item[$\ddagger$] In RSBH, $d$ denotes the number of bits in a message block.
      \item[$\sharp$] In MajorMark$^+$, $r$ with $r < b$ denotes the number of message blocks.
        \vspace{0.1em}
    \end{tablenotes}
  \end{threeparttable}

  }
\end{wraptable}
\textbf{Multi-Bit Watermarking in LLMs.}
Multi-bit watermarking aims to encode and decode a $b$-bit binary message $\mathbf{m} \in \{0,1\}^b$ (we assume that $b$ is even in this work) into and from the generated text. 
Multi-bit messages can carry richer information, such as the user’s identity, the model version, or a timestamp. To embed $\mathbf{m}$ into LLM-generated text, CTWL~\citep{ctwl} and CycleShift~\citep{cycleshift} leverage $\mathbf{m}$ to calculate the pseudo-random seed during encoding. However, their decoding processes require brute-force enumeration over all $2^b$ possible message values to find the best candidate, rendering them computationally infeasible for large $b$. A more efficient method, MPAC~\citep{mpac}, divides the message $\mathbf{m}$ into $r$ blocks $\{\mathbf{m}_1, \mathbf{m}_2, \dots, \mathbf{m}_{r}\}$, where each block is $d$ bits long. During generation, one message block is encoded per token generation step. Specifically, for each block $\mathbf{m}_i$, the encoder partitions the vocabulary into $2^d$ disjoint shards and selects the $[\mathbf{m}_i]_{10}$-th shard as the green list for that step, where $[\cdot]_{10}$ denotes the decimal (base-10) representation of the binary block. For decoding, MPAC identifies the shard with the highest token count at each step, interprets its index as the base-10 value of the corresponding message block, and converts it back to a $d$-bit binary string. The full message is then reconstructed by concatenating all recovered blocks. However, MPAC fixes $\gamma = 1/2^d$ (e.g., $\gamma = 0.25$ for $d = 2$ in their default setting), which significantly degrades the utility of the generated text. To address this, RSBH~\citep{rsbh} improves MPAC by computing hash seeds using $[\mathbf{m}_i]_{10}$, choosing a larger green list size of $\gamma = 0.5$. This enhancement improves utility but increases decoding complexity by a factor of $2^d$. 
We summarize the decoding complexity of representative multi-bit methods in Table~\ref{tab:summarize_method}.

\section{Problem Formulation and Motivation}
\label{sec: problem_and_motivation}

\textbf{Problem Formulation.}
Given an LLM $f$, a user prompt $\mathbf{x}_p$, and a $b$-bit binary message $\mathbf{m} \in \{0,1\}^b$, the goal of multi-bit watermarking is to embed $\mathbf{m}$ into the generated text to produce a watermarked output $\mathbf{x}_g^\prime$. This process is performed by an encoder function $\mathtt{Enc}(\cdot)$: $\mathbf{x}_g^\prime = \mathtt{Enc}(f, \mathbf{x}_p, \mathbf{m}),$
where $\mathbf{x}_g^\prime$ should remain fluent and semantically consistent with the unwatermarked output $\mathbf{x}_g$.  
The embedded message can be extracted using a decoder $\mathtt{Dec}(\cdot)$ to obtain
$\mathbf{m}^\prime = \mathtt{Dec}(\mathbf{x}_g^\prime)$. Formally, the objective of multi-bit watermarking can be formulated as the following constrained optimization problem:
\begin{align*}
    \max \quad & \Pr\!\Big[\mathbf{m}^\prime = \mathbf{m} \ \big|\ \mathbf{m}^\prime = 
    \mathtt{Dec}(\mathtt{Enc}(f,\mathbf{x}_p,\mathbf{m})) \Big] \\
    \text{s.t.} \quad & \mathtt{Quality}(\mathbf{x}_g^\prime \mid \mathbf{x}_p) \geq \tau,
\end{align*}
where $\mathtt{Quality}(\cdot)$ quantifies the quality of generated text, and $\tau$ is a minimum quality threshold (often chosen to be close to the quality of $\mathbf{x}_g$).  
In practice, conditional perplexity $\mathtt{PPL}(\cdot)$ is widely used as a quality metric.  
This formulation captures the key goal of watermarking: to maximize the probability of exact message decoding while ensuring negligible degradation in text quality.

\textbf{Motivation.}
Existing perturbation-based methods typically necessitate careful selection of the green list ratio $\gamma$ to balance the trade-off between the quality of watermarked text and the accuracy of decoding. 
Specifically, due to the shift invariance of the $\mathtt{softmax}(\cdot)$ function, adding the bias to a larger portion of logits (i.e., using a larger $\gamma$) better preserves the original token probability distribution, thereby maintaining the quality of the generated sequence. However, since these methods rely on counting the frequency of generated tokens that fall within the recovered green list $\mathcal{G}$, a larger $\gamma$ weakens the watermark signal, reducing the distinctiveness of the perturbation and making decoding less accurate. Conversely, using a smaller $\gamma$ introduces stronger distortions to the output distribution, thereby improving the ability to decode the embedded message. Yet this comes at the cost of generation quality: high-probability (i.e., desirable) tokens may fall outside $\mathcal{G}$ and be suppressed due to the bias, while low-probability tokens within $\mathcal{G}$ may be unintentionally amplified and incorrectly sampled. This imbalance can significantly degrade the quality of $\mathbf{x}_g^\prime$.

\setlength{\intextsep}{0mm}%
\begin{wrapfigure}{r}{0.5\textwidth} 
    \centering
    \includegraphics[width=\linewidth]{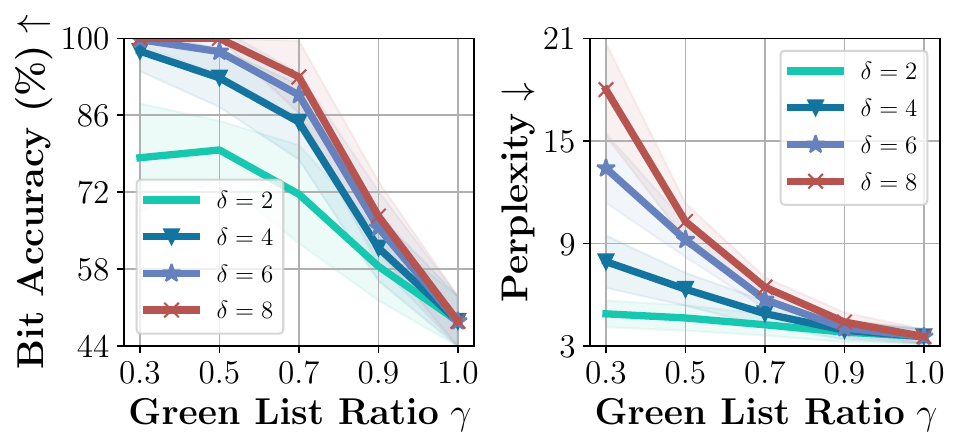}
    \caption{
    Impact of green list ratio $\gamma$ on the utility of watermarked text (right) and the decoding accuracy (left).
    }
    \label{fig:intuition}
\end{wrapfigure}

To empirically illustrate the role of $\gamma$ in shaping this trade-off, we conduct experiments using RSBH~\citep{rsbh} with LLaMA-$2$-$7$B~\citep{llama2}, varying the green list ratio $\gamma$ from $0.3$ to $1.0$ under message length $b = 64$. Note that $\gamma = 1.0$ corresponds to the extreme case where watermarking is ineffective. We report both the bit accuracy (i.e., the proportion of bits correctly decoded from the watermarked text) and the perplexity of the watermarked text in Figure~\ref{fig:intuition}, under different bias settings. As expected, increasing $\gamma$ consistently improves text quality, as reflected by lower perplexity across all bias levels, but at the cost of reduced bit accuracy. These results empirically demonstrate the trade-off between text utility and decoding accuracy influenced by $\gamma$. 
We summarize the green list ratios $\gamma$ used in prior methods in Table~\ref{tab:summarize_method}. All of them set $\gamma \leq 0.5$, by following previous works~\citep{ctwl, cycleshift, depthw, rsbh} or empirical tuning~\citep{mpac}.  

\section{Proposed Methods}

\subsection{Majority Bit-Aware Encoding}

We first formally define the majority bit $\lambda$ of a multi-bit message in Definition~\ref{def:majority_bit}.

\begin{definition}[Majority Bit]\label{def:majority_bit}
Given a binary message $\mathbf{m} \in \{0,1\}^{b}$ with $b\geq2$, let $h_0$ and $h_1$ denote the number of occurrences of bit $0$ and $1$ 
in $\mathbf{m}$, respectively. The \textit{majority bit} $\lambda \in \{0,1\}$ of $\mathbf{m}$ is defined as $\lambda=1$ if $h_1 \geq h_0$, and $\lambda=0$ otherwise. 
\end{definition}
For example, given a message $\mathbf{m}=110111 \in \{0,1\}^6 $, the majority bit is $\lambda=1$ because $h_1=5$ and $h_0=1$. In cases where the occurrences are equal (i.e., $h_0=h_1=b/2$), we define the majority bit as $\lambda=1$ by default. Crucially, the majority bit for any message appears at least $\lceil b/2 \rceil$ times, which provides a stable heuristic for message encoding.

\begin{figure}[t]
    \centering
    \begin{subfigure}[t]{0.48\textwidth} 
        \centering
        \includegraphics[width=\linewidth]{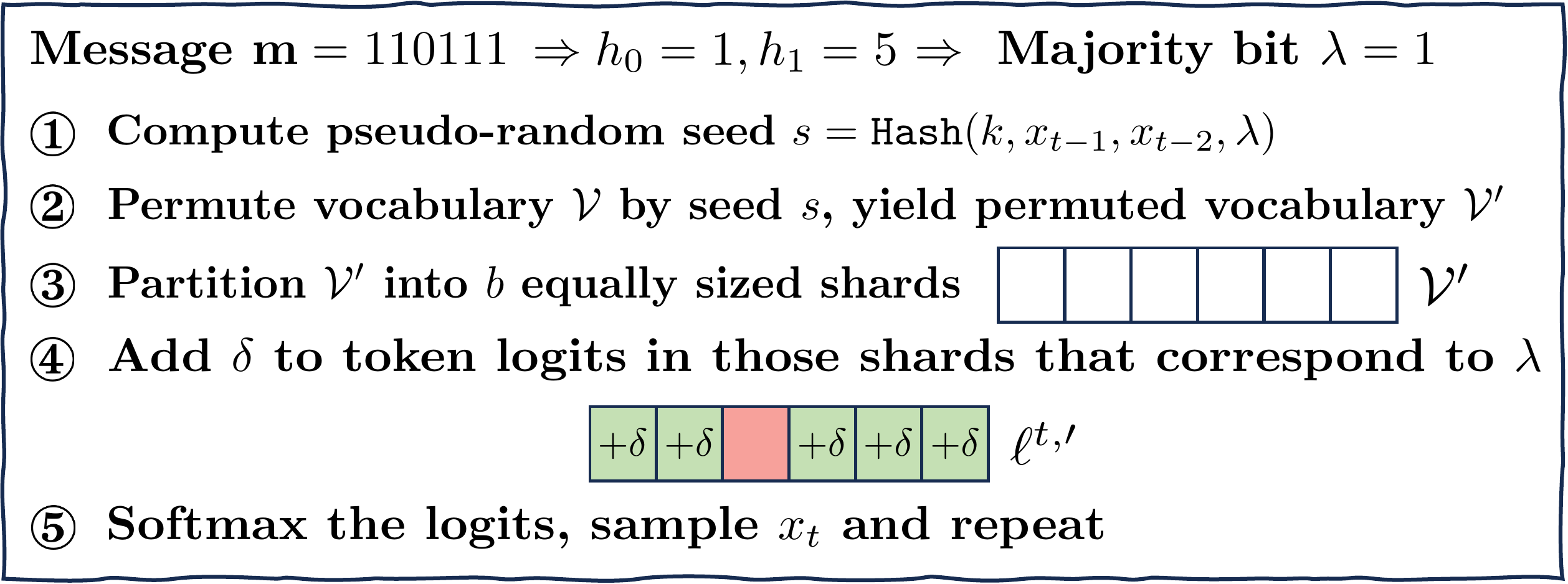}
        \caption{Majority bit-aware encoding process at generation step $t$.}
        \label{fig:majormark}
    \end{subfigure}
    \hfill
    \begin{subfigure}[t]{0.48\textwidth}
        \centering
        \includegraphics[width=\linewidth]{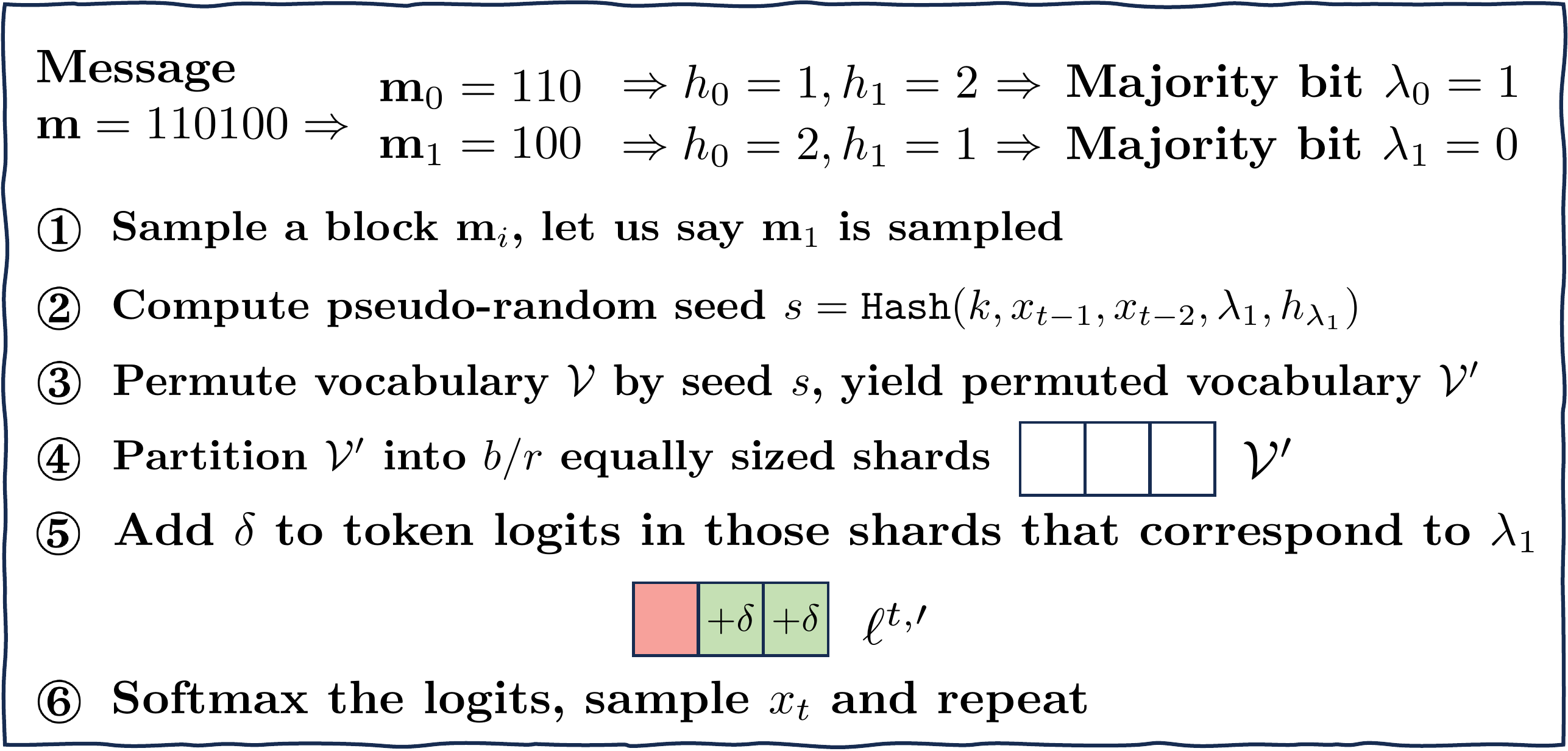}
        \caption{Block-wise majority bit-aware encoding process at generation step $t$.}
        \label{fig:majormarkplus}
    \end{subfigure}

    \caption{An overview of the majority bit-aware encoding process of MajorMark and MajorMark$^+$. 
    }
    \label{fig:majormark_combined}
\end{figure}

We present the overview of the majority bit-aware encoding process at the $t$-th token generation step of LLM in Figure~\ref{fig:majormark}. Specifically, prior to the generation process of the LLM $f$, it first identifies the majority bit $\lambda$ of the $b$-bit binary message $\mathbf{m}$ to be embedded. During the generation of the $t$-th token, after obtaining the logits $\ell^t$ from $f$, it performs the following steps:
\ding{172} Computes a pseudo-random seed $s$ based on a hash function (the design of this hash function is detailed later);
\ding{173} Applies the seed $s$ to permute the vocabulary $\mathcal{V}$, yielding a permuted vocabulary $\mathcal{V}^\prime$;
\ding{174} Evenly partitions $\mathcal{V}^\prime$ into $b$ disjoint shards $[\mathcal{V}_1, \mathcal{V}_2, \dots, \mathcal{V}_b]$, where each shard $\mathcal{V}_i$ corresponds to the $i$-th bit of the message $\mathbf{m}$. The green list $\mathcal{G}$ is constructed by taking the union of the shards associated with the majority bit value $\lambda$, such that $\mathcal{G} = \cup_{i:m_i=\lambda}\mathcal{V}_i$.
\ding{175} Uses the same seed $s$ to permute $\ell^t$, yielding permuted logits $\ell^{t, \prime}$ and adds a bias $\delta$ to the logits of tokens in $\mathcal{G}$;
\ding{176} Applies the $\mathtt{softmax}(\cdot)$ function to the biased logits, yielding a new probability distribution from which the current token $x_t$ is sampled. 
Our majority bit-aware encoding strategy ensures a minimum green list ratio of $\gamma\geq 0.5$, thereby preserving the utility of watermarked text.

\subsection{MajorMark}
We propose MajorMark, which directly utilizes majority bit-aware encoding as the encoder and is equipped with a clustering decoder. We first derive the theoretical guarantee for $\gamma$ of MajorMark. 
\begin{theorem}[$\gamma$ of MajorMark]\label{them:famark_gls}
For any message $\mathbf{m} \in \{0,1\}^{b\geq2}$, the green list $\mathcal{G}$ constructed via MajorMark satisfies $|\mathcal{G}| \geq 0.5 |\mathcal{V}|$, i.e., $\gamma \geq 0.5$. Moreover, for random $b$-bit binary messages, 
we have 
$\mathbb{E}_{\mathbf{m}}[\gamma] \approx 0.5 + 1/\sqrt{2 \pi b}$.
\begin{proof}
    The detailed proof is in Appendix~\ref{apdx: proof_of_green_list}.
\end{proof}
\begin{remark}\label{remark_gls}
MajorMark achieves a strictly larger expected $\gamma$ ($\mathbb{E}_{\mathbf{m}}[\gamma]>0.5$) than existing methods for any finite $b$, indicating improved text utility. As $b \to \infty$, the expectation $\mathbb{E}_{\mathbf{m}}[\gamma]$ gradually converges to $0.5$, making the improvements less significant. 
Note that we derive the expected $\gamma$ by assuming each bit is an independent random variable following a  $\mathrm{Bernoulli}(0.5)$ distribution. However, there are two extreme cases $\mathbf{m} = \mathbf{0}^b$ and $\mathbf{m} = \mathbf{1}^b$ that will result in $\gamma=1.0$, thereby rendering the watermarking ineffective.  
In practice, these two special cases can be explicitly disallowed during encoding, and this has a negligible impact on the expected value of $\gamma$ when $b$ is large. 
\end{remark}
\end{theorem}

\textbf{Design of the Hash Function.}
We revisit the design of the hash function in majority bit-aware encoding. Prior works~\citep{kgw, mpac} typically compute the seed $s$ as $\mathtt{Hash}(k, x_{t-1})$, where $k$ is a secret key and $x_{t-1}$ is the previous token. However, frequent words (e.g., \textit{the}, \textit{is}, \textit{to}) appear disproportionately in generated text, causing certain seeds, and thus token-to-shard mappings, to repeat~\citep{rsbh}. This leads to repetitive green lists $\mathcal{G}$, over-sampling of specific tokens, and degraded text quality. To address this, we follow~\citep{depthw} and extend the hash input to include the second-to-last token $x_{t-2}$, using $\mathtt{Hash}(k, x_{t-1}, x_{t-2})$. This yields more diverse token-to-shard mappings. We further incorporate the majority bit $\lambda$, forming $\mathtt{Hash}(k, x_{t-1}, x_{t-2}, \lambda)$, which increases seed diversity and facilitates recovery of $\lambda$ during decoding, as detailed below.

\textbf{Clustering-Based Decoding.} The message decoding process takes as input a generated text $\mathbf{x}$ of length $T$ and produces the decoded message $\mathbf{m}^\prime$. Specifically, for each token $x_t \in \mathbf{x}$ with $t = 3, 4, \dots, T$, MajorMark reconstructs the token-to-shard mapping for $x_t$ used in encoding. We discard the first two tokens $x_1$ and $x_2$ from decoding, as computing their token-to-shard mappings requires access to the prompt tokens in $\mathbf{x}_p$, which are unavailable during decoding. To reconstruct the token-to-shard mapping for $x_t$, we first compute the seed $s$ using $\mathtt{Hash}(k, x_{t-1}, x_{t-2}, \lambda)$. We then determine the shard to which $x_t$ belongs and increment its corresponding count. After processing all tokens, we obtain a shard-wise token occurrence count that reflects the frequency of tokens belonging to each shard. Based on this occurrence count, we apply a clustering algorithm (e.g., $\mathtt{KMeans}$~\citep{kmeans} with $K=2$) to partition the shards into two clusters. The cluster with the higher average token count, denoted by $\mathcal{C}$, is interpreted as corresponding to the majority bit $\lambda$, as the logits of tokens in these shards were positively perturbed by $\delta$ during encoding. The message bits are then recovered by setting $m_i = \lambda$ for each shard $i \in \mathcal{C}$ and $m_i = 1 - \lambda$ for those not in the cluster (i.e., $i \notin \mathcal{C}$). We provide the theoretical justification of this decoding and analyze decoding errors in Appendix~\ref{apdx: theo_cluster}.

\setlength{\intextsep}{0mm}%
\begin{wrapfigure}{r}{0.45\textwidth}
    \centering
    \includegraphics[width=\linewidth]{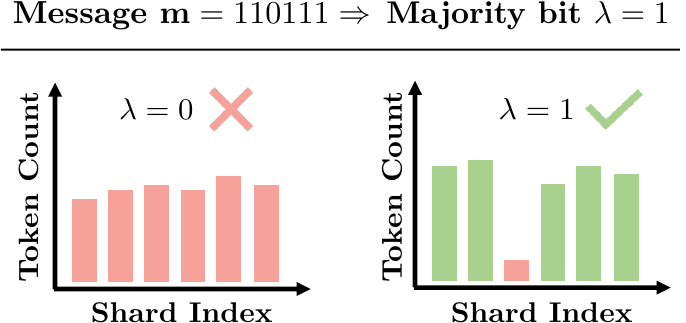}
\caption{Occurrence distributions for candidate $\lambda$ (0 or 1). Distribution produced by the correct $\lambda$ exhibits large dispersion.}
    \label{fig: decoding}
\end{wrapfigure}

One remaining challenge in MajorMark's decoding process is determining the correct value of $\lambda$. Recall that we explicitly include $\lambda$ as the input of the hash function $\mathtt{Hash}(k, x_{t-1}, x_{t-2}, \lambda)$. This design allows the recovery of $\lambda$. The key intuition is that only the correct seed $s$ can reproduce the token-to-shard mappings used during encoding. Therefore, in the decoding phase, we compute the shard-wise token occurrence counts for each possible value of $\lambda$ (i.e., $0$ or $1$). The correct value of $\lambda$ is determined as the one that produces a more dispersed occurrence distribution (e.g., exhibiting a larger standard deviation), whereas the incorrect value yields a nearly uniform distribution, as illustrated in Figure~\ref{fig: decoding}. This approach enables reliable recovery of the embedded message with only a one-time additional enumeration over $\mathbf{x}$. 
The detailed encoding and decoding algorithms of MajorMark and the implementation of the hash function are presented in Appendix~\ref{sec: algorithm_of_majormark}.

\subsection[\texorpdfstring{MajorMark+: An Improved Method}{MajorMark+: An Improved Method}]
        {MajorMark$^+$: An Improved Method}

We now introduce MajorMark$^+$, an enhanced version of MajorMark that further improves both the quality of watermarked text and decoding accuracy. Specifically, compared with MajorMark, MajorMark$^+$ (1) guarantees a larger expected $\gamma$ through a \textit{block-wise majority bit-aware encoding} strategy; and (2) enhances decoding accuracy by replacing the boundary-sensitive clustering-based decoding with a \textit{deterministic decoding} method.

\textbf{Block-Wise Majority Bit-Aware Encoding.}
As we discussed in Remark~\ref{remark_gls}, when $b$ becomes large, the improvement of $\gamma$ achieved by MajorMark becomes less significant as $\gamma$ converges to $0.5$. Inspired by previous methods that divide the full message into several blocks~\citep{mpac, rsbh, stealthink} to perform watermark encoding, MajorMark$^+$ is equipped with a \textit{block-wise majority bit-aware encoding} method to slow the convergence of $\gamma$ given a finite $b$, especially when $b$ is large. We present an overview of MajorMark$^+$'s block-wise majority bit-aware encoding process at generation step $t$ in Figure~\ref{fig:majormarkplus}. Specifically, given any message $\mathbf{m} \in \{0,1\}^{b}$ with $b \geq 2$, MajorMark$^+$ divides it into $r$ equal-sized blocks $\{ \mathbf{m}_1, \mathbf{m}_2, \dots, \mathbf{m}_r \}$, where $r$ divides $b$. During the $t$-th generation step of the LLM $f$, MajorMark$^+$ pseudo-randomly assigns the $i$-th block $\mathbf{m}_i$ to the current token $x_t$, where $i = (x_{t-2} + x_{t-1}) \mod r$. As a result, each token contributes to encoding only one message block.

To encode a given block $\mathbf{m}_i$ where $i \in [r]$, MajorMark$^+$ applies the same \textit{majority bit-aware encoding} strategy used in MajorMark: the green list is constructed based on the block-specific majority bit $\lambda_i$, and the logits are perturbed accordingly. Importantly, the hash function of MajorMark$^+$ is in the form of $\mathtt{Hash}(k, x_{t-1}, x_{t-2}, \lambda_i, h_{\lambda_i})$. Specifically, we further incorporate the number of occurrences of the majority bit $h_{\lambda_i}$ into the calculation of the hash seed $s$, which enables deterministic decoding in MajorMark$^+$, as discussed later. We now present the formal guarantee on MajorMark$^+$'s expected $\gamma$ as follows in Theorem~\ref{them:majormarkplus_gls}.

\begin{theorem}[$\gamma$ of MajorMark$^+$]\label{them:majormarkplus_gls}
MajorMark$^+$ preserves the lower bound of the green list ratio as in MajorMark, i.e., $|\mathcal{G}| \geq 0.5 |\mathcal{V}|$ and $\gamma \geq 0.5$. Moreover, for random $b$-bit binary messages, we have $
\mathbb{E}_{\mathbf{m}}[\gamma] \approx 0.5 + 1/\sqrt{2 \pi (b/r)} .$
\begin{proof}
    The detailed proof is in Appendix~\ref{apdx: proof_of_green_list_famarkplus}.
\end{proof}
\begin{remark}

Given any finite $b$, MajorMark$^+$ always guarantees a larger expected $\gamma$ than MajorMark for any $r >1$. This property improves the text quality of MajorMark$^+$. Recall from Remark~\ref{remark_gls} that MajorMark discards extreme cases where the message is entirely composed of zeros or ones, i.e., $\mathbf{m} = \mathbf{0}^b$ or $\mathbf{m} = \mathbf{1}^b$. MajorMark$^+$ similarly excludes such extreme messages at the block level. \textit{These infeasible cases are extremely rare and do not meaningfully affect the practical effectiveness of MajorMark$^+$ when $b$ is large.} Specifically, the number of infeasible codes is given by $2^b - (2^{b/r} - 2)^r$. For instance, given a large $b = 32$ and $r = 2$, the total codes are $2^{32}$, while the number of infeasible codes is only $262{,}140$, accounting for approximately $0.006\%$ of the code space.
\end{remark}
\end{theorem}

\textbf{Deterministic Decoding.}
When the message length $b$ is large and the watermarking bias $\delta$ is not large enough, the resulting shard-wise token occurrence count becomes less dispersed. This weak signal may degrade the effectiveness of MajorMark’s decoding process in specific cases, as clustering algorithms are \textit{unsupervised} and inherently sensitive to data points near \textit{decision boundaries}, failing to accurately separate the shards. To address this limitation, MajorMark$^+$ introduces a deterministic decoding method designed to recover the majority bit $\lambda_i$ and its occurrences $h_{\lambda_i}$ in the message for each block with high reliability. Specifically, given the hash function $\mathtt{Hash}(k, x_{t-1}, x_{t-2}, \lambda_i, h_{\lambda_i})$, MajorMark$^+$ exhaustively enumerates all possible combinations of $\lambda_i$ and $h_{\lambda_i}$. For each combination, it reconstructs the corresponding shard-wise token occurrence count. The correct configuration yields a distinctly dispersed occurrence counts, while incorrect configurations result in near-uniform counts. By identifying the configuration that induces the largest dispersion, MajorMark$^+$ reliably recovers the correct values of $\lambda_i$ and $h_{\lambda_i}$. 
The message bits corresponding to those shards with top-$h_{\lambda_i}$ token occurrences are then assigned the recovered bit $\lambda_i$, while the remaining bits are assigned $1 - \lambda_i$. 
%
%
The full message is subsequently reconstructed by assembling all blocks. This deterministic approach eliminates the reliance on clustering, thereby avoiding sensitivity to ambiguous token counts and improving decoding accuracy. The detailed encoding and decoding algorithms of MajorMark$^+$ and the implementation of the hash function are presented in Appendix~\ref{sec: algorithm_of_majormarkplus}.


\textbf{Decoding Complexity Analysis.}
MajorMark$^+$ needs to identify both the majority bit $\lambda_i$ and its occurrence $h_{\lambda_i}$ for each block. When block-level all-zero and all-one codes are excluded, each block has $b/r-1$ feasible configurations. Since the block assignment of each token is independent of the hypothesized $(\lambda_i,h_{\lambda_i})$, the decoder can evaluate the corresponding configurations for all blocks within the same pass over the unverified token sequence $\mathbf{x}$. Therefore, MajorMark$^+$ requires only $(b/r-1)\times|\mathbf{x}|$ decoding operations, which is much lower than enumeration-based methods such as CycleShift~\citep{cycleshift} and DepthW~\citep{depthw}, which require $2^b$ enumerations.

\section{Experiments}
\subsection{Experimental Settings}
\label{sec: exp_settings}
\textbf{General Settings.}
By default, we consider a total of $20$ users, corresponding to $20$ randomly generated messages. Each user submits two prompts, and each prompt is used to generate a text of $T/2 = 250$ tokens, resulting in $T = 500$ tokens per user for decoding. We compare our proposed methods with four state-of-the-art perturbation-based methods, including DepthW~\citep{depthw}, CycleShift~\citep{cycleshift}, MPAC~\citep{mpac}, and RSBH~\citep{rsbh}. In the larger-scale evaluation, we additionally compare against the distortion-free methods StealthInk~\citep{stealthink} and BiMark~\citep{feng2025bimark}. For DepthW and CycleShift, we only evaluate the case of $b = 8$, as their decoding time grows exponentially and becomes impractical for larger message lengths. For MajorMark, we use $\mathtt{KMeans}$~\citep{kmeans} clustering with $K = 2$ for decoding, and all other hyperparameters of $\mathtt{KMeans}$ are kept at their default settings. For MajorMark$^+$, the default number of blocks $r$ is set to $2$. All experiments were carried out on a self-managed Linux-based computing cluster running Ubuntu 20.04.6 LTS. The cluster is equipped with eight NVIDIA RTX A6000 GPUs (each with 48 GB of memory) and AMD EPYC 7763 CPUs featuring 64 cores. In total, the experiments accumulated roughly two weeks of GPU compute time.

\textbf{Datasets and Models.}
Following prior work~\citep{kgw, mpac, rsbh}, we primarily evaluate our method on the text completion task using the C$4$ news dataset~\citep{c4}. Unless otherwise stated, we use LLaMA-2-7B~\citep{llama2}. We additionally evaluate recent public LLMs in Section~\ref{sec: results}, with further results on additional models, datasets, and benchmark tasks provided in Appendix~\ref{apdx: additional_results}.

\textbf{Metrics.}
We evaluate the quality of generated text using perplexity (PPL) computed by a larger LLaMA-$2$-$13$B model. In addition, we report BERTScore and ROUGE in Appendix~\ref{apdx: semantic_fidelity}, where the results show that our method maintains higher semantic and lexical quality than the other methods. We also report the Top-$5$ hit rate, which measures the proportion of generated tokens that appear among the top-$5$ logits of the original model output, indicating how often desirable tokens are still sampled.  For decoding accuracy, we adopt bit accuracy (BA), following prior works~\citep{ba1,ba2,ba3,mpac}, defined as the proportion of correctly decoded bits relative to the ground-truth message. 

\begin{table*}[t]
  \centering
  \caption{
Comparison of methods across various message lengths and watermarking biases. The PPL of non-watermarked text is $3.81$, serving as a lower bound for the PPL of all watermarked texts.
}
  \scalebox{0.75}{
    \begin{tabular}{l|ccc|ccc|ccc|cc}
    \toprule
    \footnotesize
    \multirow{2}[2]{*}{\textbf{Method}} & \multicolumn{3}{c|}{$\delta=2$}   & \multicolumn{3}{c|}{$\delta=4$}   & \multicolumn{3}{c|}{$\delta=6$} & \multirow{2}[2]{*}{\textbf{\makecell*[c]{Avg. \\ BA$\uparrow$}}} & \multirow{2}[2]{*}{\textbf{\makecell*[c]{Avg. \\ PPL$\downarrow$}}} \\
    \cmidrule(r){2-4}\cmidrule(r){5-7}\cmidrule(r){8-10}
          & \textbf{BA}$\uparrow$    & \textbf{PPL}$\downarrow$   & \textbf{Top-$\mathbf{5}$}$\uparrow$ & \textbf{BA}$\uparrow$    & \textbf{PPL}$\downarrow$   & \textbf{Top-$\mathbf{5}$}$\uparrow$ & \textbf{BA}$\uparrow$    & \textbf{PPL}$\downarrow$   & \textbf{Top-$\mathbf{5}$}$\uparrow$ \\
    \midrule
    \multicolumn{10}{l}{\textit{Message length $b=8$}} \\
    CycleShift & $\mathbf{100.00}$ & $5.09$ & $90.72$ & $\mathbf{100.00}$ & $8.64$ & $81.43$ & $\mathbf{100.00}$ & $15.44$ & $75.65$ & $\mathbf{100.00}$ & $9.72$ \\
    DepthW & $95.62$ & $4.53$ & $91.65$ & $\mathbf{100.00}$ & $8.24$ & $79.53$ & $\mathbf{100.00}$ & $25.49$ & $61.13$ & $98.54$ & $12.75$ \\
    MPAC  & $99.38$ & $5.03$ & $90.41$ & $\mathbf{100.00}$ & $8.74$ & $81.44$ & $\mathbf{100.00}$ & $16.06$ & $74.52$ & $99.79$ & $9.94$ \\
    RSBH  & $\mathbf{100.00}$ & $4.80$ & $91.19$ & $\mathbf{100.00}$ & $6.36$ & $89.30$ & $\mathbf{100.00}$ & $8.66$ & $86.87$ & $\mathbf{100.00}$ & $6.61$ \\
    \rowcolor[rgb]{ .9, .9, .9}MajorMark & $\mathbf{100.00}$ & $4.50$ & $92.19$ & $\mathbf{100.00}$ & $5.87$ & $\mathbf{90.65}$ & $\mathbf{100.00}$ & $7.81$ & $89.23$ & $\mathbf{100.00}$ & $6.06$ \\
    \rowcolor[rgb]{ .9, .9, .9}MajorMark$^+$ & $\mathbf{100.00}$ & $\mathbf{4.43}$ & $\mathbf{92.43}$ & $\mathbf{100.00}$ & $\mathbf{5.75}$ & $90.36$ & $\mathbf{100.00}$ & $\mathbf{6.97}$ & $\mathbf{90.06}$ & $\mathbf{100.00}$ & $\mathbf{5.72}$ \\
    \midrule
    \multicolumn{10}{l}{\textit{Message length $b=32$}} \\
    MPAC  & $89.22$ & $5.03$  & $90.09$ & $98.75$ & $9.37$  & $80.75$ & $\mathbf{100.00}$   & $15.32$ & $75.35$ &$95.99$ & $9.91$  \\
    RSBH  & $89.38$ & $4.68$  & $92.15$ & $97.66$ & $6.43$  & $88.48$ & $97.50$  & $8.47$  & $87.13$ &$94.85$ & $6.53$  \\
    \rowcolor[rgb]{ .9, .9, .9}MajorMark 
          & $96.74$ & $\mathbf{4.43}$  & $91.90$  
          & $99.06$ & $6.32$  & $89.28$ 
          & $\mathbf{100.00}$   & $8.36$  & $88.09$  & $98.60$ & $6.37$ \\
    \rowcolor[rgb]{ .9, .9, .9}MajorMark$^+$ 
          & $\mathbf{97.81}$ & $4.49$  & $\mathbf{92.42}$ 
          & $\mathbf{100.00}$   & $\mathbf{6.05}$  & $\mathbf{90.17}$ 
          & $\mathbf{100.00}$   & $\mathbf{7.90}$   & $\mathbf{88.49}$ & $\mathbf{99.27}$ & $\mathbf{6.15}$ \\
    \midrule
    \multicolumn{10}{l}{\textit{Message length $b=64$}} \\
    MPAC  & $81.48$ & $4.99$ & $90.08$ & $93.59$ & $8.64$ & $82.01$ & $96.48$ & $15.95$ & $75.14$ & $90.52$ & $9.86$ \\
    RSBH  & $81.25$ & $4.93$ & $91.80$ & $90.39$ & $6.41$ & $88.91$ & $97.58$ & $9.22$  & $87.06$ & $89.74$ & $6.85$ \\
    \rowcolor[rgb]{ .9, .9, .9}MajorMark   
          & $83.59$ & $4.78$ & $91.46$ 
          & $93.83$ & $\mathbf{6.11}$ & $\mathbf{89.34}$ 
          & $98.67$ & $8.82$ & $87.27$ & $92.03$ & $6.57$  \\
    \rowcolor[rgb]{ .9, .9, .9}MajorMark$^+$   
          & $\mathbf{86.95}$ & $\mathbf{4.74}$ & $\mathbf{91.83}$ 
          & $\mathbf{97.81}$ & $6.21$ & $89.30$ 
          & $\mathbf{99.69}$ & $\mathbf{7.93}$ & $\mathbf{88.43}$ & $\mathbf{94.82}$ & $\mathbf{6.29}$ \\
    \bottomrule
    \end{tabular}%
    }
  \label{tab:main_result_table}%
  \vspace{-10pt}
\end{table*}

\subsection{Empirical Results} \label{sec: results}

\textbf{Main Results.}
We conduct a comprehensive evaluation of representative perturbation-based methods under varying message lengths ($b \in \{8, 32, 64\}$) and watermarking biases ($\delta \in \{2, 4, 6\}$). Table~\ref{tab:main_result_table} reports BA, PPL, and Top-$5$ hit ratio for each setting. When $b = 8$, all methods achieve high BA across all watermarking bias settings. In this low-bit regime, MajorMark and MajorMark$^+$ exhibit the best text quality, reflected by their average PPLs of $6.06$ and $5.72$, which are the second-lowest and lowest among all methods. As $b$ increases, all methods generally exhibit lower BA and higher PPL due to the increased difficulty of encoding and decoding longer messages. Nonetheless, MajorMark and MajorMark$^+$ consistently maintain strong performance, outperforming MPAC and RSBH in both PPL and BA. For example, when $b = 64$, MajorMark and MajorMark$^+$ achieve PPLs of $6.57$ and $6.29$, which represent improvements of $+0.28$ and $+0.56$ over RSBH's PPL of $6.85$. These improvements stem from the majority bit-aware encoding strategy, which enables a larger green list during encoding, thereby better preserving the text quality. Furthermore, our methods achieve higher Top-$5$ hit rates across all settings, further demonstrating their ability to maintain text quality. In terms of BA, they achieve $92.03\%$ and $94.82\%$, outperforming MPAC's $90.52\%$ by $+1.51\%$ and $+4.30\%$. 
We report the standard deviations and 95\% CIs for the results with $b=\{32, 64\}$ in Appendix~\ref{app:full_main_table}.

\setlength{\intextsep}{0mm}%
\begin{wrapfigure}{r}{0.51\textwidth}
    \centering
    \includegraphics[width=\linewidth]{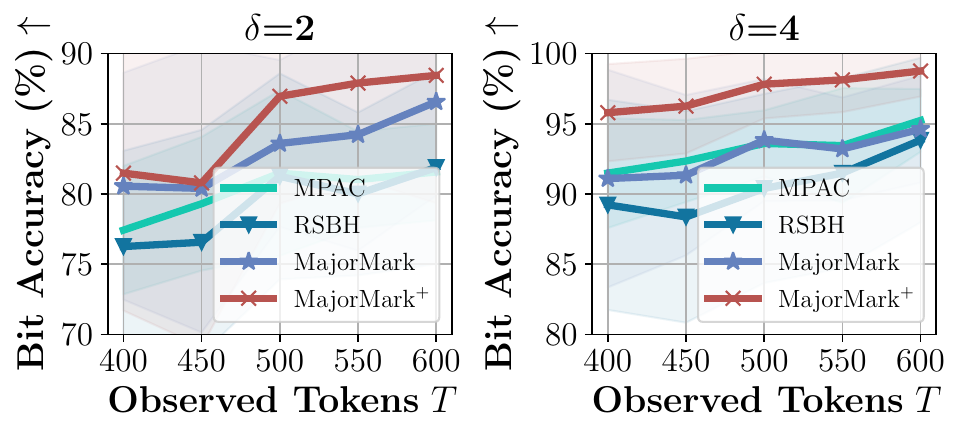}
    \caption{
BA of different methods under varying numbers of observed tokens $T$ with $b = 64$.
}

    \label{fig:AvaTokens}
\end{wrapfigure}

\textbf{Results on Available Tokens $T$ for Decoding.} Figure~\ref{fig:AvaTokens} reports BA under different numbers of observed tokens $T \in \{400, 450, 500, 550, 600\}$, with $b=64$ and $\delta=2$ (left) and $\delta=4$ (right). As expected, BA increases for all methods as $T$ grows. Under $\delta=2$, MajorMark and MajorMark$^+$ consistently outperform MPAC and RSBH across all token budgets, indicating higher decoding efficiency when only limited tokens are available. When $\delta$ increases to $4$, the advantage of MajorMark$^+$ becomes more pronounced. For example, at $T=400$, MPAC and RSBH achieve BA of $91.48\%$ and $89.22\%$, respectively, while MajorMark$^+$ reaches $95.78\%$, improving BA by $+6.56\%$ over MPAC and $+4.30\%$ over RSBH. 

\setlength{\intextsep}{0mm}%
\begin{wrapfigure}{r}{0.51\textwidth}
    \centering
    \includegraphics[width=\linewidth]{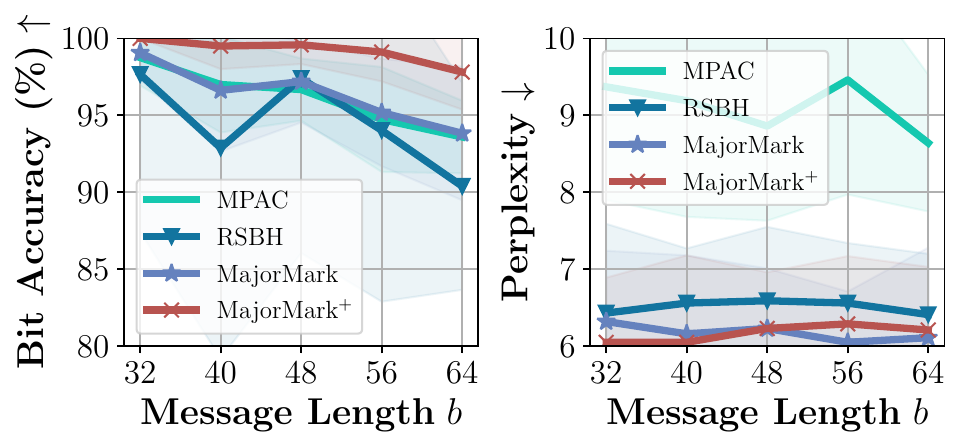}
    \caption{BA and PPL of methods with increasing message length $b \in \{32, 40, 48, 56, 64\}$ with $\delta=4$.
}
    \label{fig:ba_ppl_with_b}
\end{wrapfigure}

\textbf{Additional Results on Message Length.} 
We further investigate how the message length ($b \in \{32, 40, 48, 56, 64\}$) affects the performance of different watermarking methods on decoding accuracy (left) and text quality (right) under fixed decoding token budget $T=500$, as shown in Figure~\ref{fig:ba_ppl_with_b}. A larger $b$ increases the complexity of the decoding process, which generally leads to reduced BA. As expected, all methods exhibit a decline in BA as $b$ increases. MajorMark maintains a comparable BA to MPAC but achieves significantly better PPL. Notably, our proposed method, MajorMark$^+$, consistently yields the highest BA while maintaining a low PPL across all message lengths, demonstrating its robustness even when $b$ is as large as $64$. These results demonstrate that under a fixed decoding token budget, our methods MajorMark and MajorMark$^+$ not only maintain high decoding accuracy but also ensuring high quality of the watermarked texts.

\setlength{\intextsep}{0mm}%
\begin{wraptable}{r}{0.45\textwidth}
    \centering
    \caption{
    Evaluation on recent public LLMs with $100$ users.
    }
    \label{tab:recent_llms}
    \resizebox{\linewidth}{!}{
    \begin{tabular}{l|cc|cc}
        \toprule
        \multirow{2}{*}{\textbf{Method}} 
        & \multicolumn{2}{c|}{\textbf{LLaMA-$3.1$-$8$B}} 
        & \multicolumn{2}{c}{\textbf{Qwen$3$-$8$B}} \\
        \cmidrule(r){2-3}\cmidrule(r){4-5}
        & \textbf{PPL}$\downarrow$ & \textbf{BA}$\uparrow$ 
        & \textbf{PPL}$\downarrow$ & \textbf{BA}$\uparrow$ \\
        \midrule
        MPAC        & $8.34$ & $97.72$ & $5.65$ & $92.66$ \\
        StealthInk  & $\mathbf{4.83}$ & $83.84$ & $\mathbf{3.50}$ & $49.31$ \\
        BiMark      & $8.39$ & $99.47$ & $5.05$ & $96.66$ \\
        \rowcolor[rgb]{0.9,.9,.9}MajorMark   & $6.79$ & $98.41$ & $4.91$ & $96.91$ \\
        \rowcolor[rgb]{0.9,.9,.9}MajorMark$^+$ & $6.51$ & $\mathbf{99.81}$ & $4.75$ & $\mathbf{98.50}$ \\
        \bottomrule
    \end{tabular}
    }
\end{wraptable}
\textbf{Larger-Scale Experiments.} To further evaluate the generality of our method, we conduct additional experiments with $100$ users on two recent public LLMs, including LLaMA-$3.1$-$8$B and Qwen$3$-$8$B~\cite{yang2025qwen3technicalreport}, and two distortion-free watermarking methods StealthInk~\cite{stealthink} and BiMark~\cite{feng2025bimark}. The results with $b=32$ and $\delta=3$ for MPAC and our methods are summarized in Table~\ref{tab:recent_llms}. Note that for StealthInk and BiMark we adopt their default settings. As shown in Table~\ref{tab:recent_llms}, StealthInk maintains relatively low PPL, but its decoding accuracy is less reliable, with BA dropping to $49.31\%$ on Qwen$3$-$8$B, which is close to random guessing. In contrast, MPAC and BiMark achieve higher BA, but they introduce larger degradation in text quality, especially on LLaMA-$3.1$-$8$B, where their PPL values exceed $8.3$. Compared with these baselines, MajorMark$^+$ achieves a better balance between text quality and decoding accuracy across both models. Specifically, it achieves the highest BA on both LLaMA-$3.1$-$8$B and Qwen$3$-$8$B, reaching $99.81\%$ and $98.50\%$, respectively, while maintaining substantially lower PPL than the high-BA baselines MPAC and BiMark.

\setlength{\intextsep}{0mm}%
\begin{wrapfigure}{r}{0.51\textwidth}
    \centering
    \includegraphics[width=\linewidth]{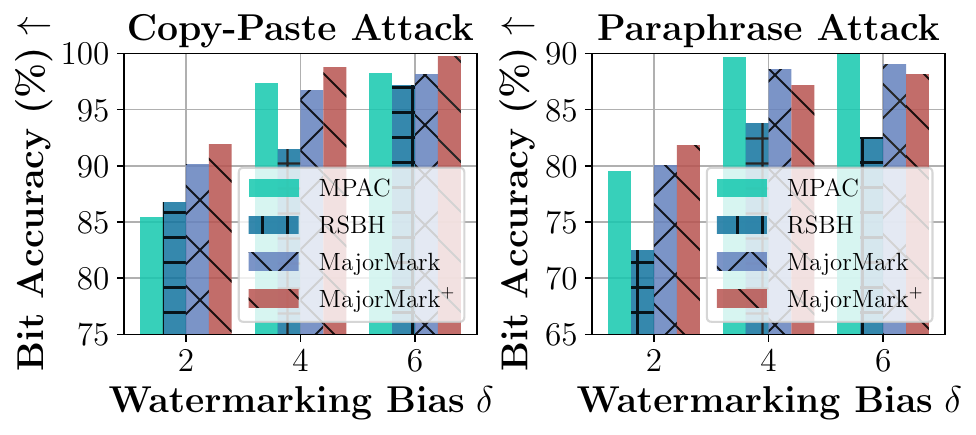}
    \caption{BA of different methods under Copy-Paste and Paraphrase attacks.}
    
    \label{fig: cp}
    \vspace{0.5em}
\end{wrapfigure}
\textbf{Robustness Against Attacks.}
We evaluate the robustness of our methods against two widely studied post-generation text editing attacks: \textit{Copy-Paste} and \textit{Paraphrase} attacks~\citep{zhang2024remark}. Following previous works~\citep{mpac, rsbh}, we replace $10\%$ of the watermarked text with non-watermarked text while keeping the total length unchanged. For the Paraphrase attack, we use the DIPPER proposed in~\citep{opengen} as the paraphraser with $L=20$ and $O=20$. Figure~\ref{fig: cp} reports BA results under both attacks, evaluated across watermarking biases $\delta \in \{2, 4, 6\}$ with a fixed message length $b=32$. For the Copy-Paste attack, MajorMark and MajorMark$^+$ generally outperform MPAC and RSBH across $\delta$ values, demonstrating their robustness even when some tokens are unwatermarked in the modified text. Paraphrase attacks remain a significant open challenge in the literature: all methods experience BA degradation under this attack. While MPAC benefits from its small $\gamma$ and achieves higher BA at larger $\delta$, our methods maintain comparable performance to MPAC and significantly outperform RSBH. Notably, MajorMark achieves slightly higher BA than MajorMark$^+$, likely due to its simpler decoding process. More specifically, MajorMark$^+$ requires enumerating all combinations of majority bits and their occurrences for each message block. Under the Paraphrase attack, token occurrences become more diverse and less reliable, leading to a slight degradation in BA of MajorMark$^+$. Additionally, we theoretically analyze the robustness of our methods under text editing attacks in Appendix~\ref{apdx: theo_robustness_para}.

\setlength{\intextsep}{0mm}%
\begin{wraptable}{r}{0.37\textwidth}
    \centering
    \caption{
    Robustness evaluation under character-level perturbation (CharP) and sentence shuffling (SenS).
    }
    \label{tab:perturbation_robustness}
    \resizebox{\linewidth}{!}{
    \begin{tabular}{l|c|cc}
        \toprule
        \multirow{2}{*}{\textbf{Method}} 
        & \multirow{2}{*}{\textbf{PPL}$\downarrow$} 
        & \multicolumn{2}{c}{\textbf{BA}$\uparrow$} \\
        \cmidrule(r){3-4}
        & & \textbf{CharP} & \textbf{SenS} \\
        \midrule
        MPAC          & $6.36$ & $81.88$ & $95.31$ \\
        RSBH          & $5.47$ & $78.28$ & $93.25$ \\
        \rowcolor[rgb]{.9,.9,.9}MajorMark     & $5.31$ & $\mathbf{82.84}$ & $94.97$ \\
        \rowcolor[rgb]{.9,.9,.9}MajorMark$^+$ & $\mathbf{5.19}$ & $81.56$ & $\mathbf{96.72}$ \\
        \bottomrule
    \end{tabular}
    }
\end{wraptable}
\textbf{Sentence Shuffling and Character-Level Perturbation Attacks.} We further evaluate the robustness of different methods under two types of text perturbations: sentence shuffling (SenS) and character-level perturbation (CharP)~\cite{zhang2025character}. For SenS, we randomly shuffle the sentences in the generated text. For CharP, following~\citep{zhang2025character}, we randomly perturb $10\%$ of the characters. The results are summarized in Table~\ref{tab:perturbation_robustness}. Since all methods rely on token-level decoding, SenS causes only limited BA degradation. In contrast, CharP substantially reduces the decoding accuracy of all methods because character-level perturbations can change tokenization and disturb the recovered green-list statistics. Despite this stronger perturbation, MajorMark and MajorMark$^+$ still achieve competitive decoding accuracy compared with MPAC and RSBH, while maintaining lower PPL. In particular, MajorMark$^+$ achieves the best BA under SenS, reaching $96.72\%$, and obtains the lowest PPL among all methods.

\setlength{\intextsep}{0mm}%
\begin{wraptable}{r}{0.5\textwidth}
\centering
\caption{Comparison of runtime (in seconds) and BA. The decoding time for DepthW is estimated.}
\label{tab:runtime_comparison}
\scalebox{0.80}{
\begin{tabular}{l|ccc}
\toprule
\textbf{Method} & \textbf{Enc.} (s) & \textbf{Dec.} (s) & \textbf{BA}$\uparrow$ \\
\midrule
DepthW & $37.87$ & $\approx 8.89 \times 10^8$ (Est.) & - \\
MPAC & $36.13$ & $0.13$ & $89.22$ \\
RSBH & $36.06$ & $25.56$ & $89.38$ \\
\rowcolor[rgb]{  .9, .9, .9}MajorMark & $36.58$ & $2.18$ & $96.74$ \\
\rowcolor[rgb]{  .9, .9, .9}MajorMark$^+$ & $36.85$ & $12.70$ & $97.81$ \\
\bottomrule
\end{tabular}
}
\end{wraptable}
\textbf{Runtime Analysis.} We evaluate the computational overhead by measuring the average encoding and decoding time under default settings ($b=32$), with results shown in Table~\ref{tab:runtime_comparison}. For encoding, our methods add negligible overhead compared to standard generation (about $36$s). For decoding, MajorMark requires two passes over the suspect text and an additional clustering step, leading to a moderate overhead compared to MPAC ($2.18$s vs. $0.13$s). MajorMark$^+$ performs $b/ r-1$ decoding passes and takes $12.70$s in total. This is still much faster than exponential-time methods such as DepthW (about $8.89 \times 10^8$s), and also faster than RSBH ($25.56$s). Meanwhile, our methods achieve substantially higher BA than existing approaches, showing that they introduce only modest computational cost while providing strong decoding performance for practical watermark verification.



\setlength{\intextsep}{0mm}%
\begin{wraptable}{r}{0.5\textwidth}
  \centering
\caption{Ablation study of MajorMark (first three rows) and MajorMark$^+$ (last four rows), evaluating clustering methods and block sizes, respectively.}
  \scalebox{0.75}{
    \begin{tabular}{l|cccc|cc}
    \toprule
    \multirow{2}[2]{*}{\textbf{Method}} & \multicolumn{2}{c}{$\delta=2$} & \multicolumn{2}{c|}{$\delta=4$}& \multirow{1}[2]{*}{\textbf{\makecell*[c]{Avg. \\ BA$\uparrow$}}} & \multirow{1}[2]{*}{\textbf{\makecell*[c]{Avg. \\ PPL$\downarrow$}}} \\
\cmidrule(r){2-3}\cmidrule(r){4-5}         & \textbf{BA}$\uparrow$    & \textbf{PPL}$\downarrow$   & \textbf{BA}$\uparrow$    & \textbf{PPL}$\downarrow$ \\
\midrule
    \rowcolor[rgb]{  .9, .9, .9}$\mathtt{KMeans}$ & $96.74$ & $4.43$  &$99.06$ & $6.32$ & $97.90$ & $5.38$\\
    $\mathtt{AC}$ & $94.06$ &  $4.43$ & $99.84$ & $6.32$ & $96.95$ & $5.38$\\
    $\mathtt{GMM}$ & $92.81$ & $4.43$ & $99.84$ &  $6.32$ & $96.33$ & $5.38$ \\
    \midrule \midrule
    $r=1$   &   $95.94$    &  $4.74$     &   $100.00$    &  $6.55$ & $97.97$ & $5.65$\\
    \rowcolor[rgb]{  .9, .9, .9}$r=2$   &  $97.81$     &    $4.49$   &   $100.00$    & $6.05$ & $98.91$ & $5.27$ \\
    $r=4$   &    $90.00$   &    $4.42$   &   $98.44$    & $5.61$ & $94.22$ & $5.02$ \\
    $r=8$   &   $91.88$    &    $4.46$   &    $98.59$   & $5.20$ & $95.24$ &$4.83$ \\
    \bottomrule
    \end{tabular}%
    }
  \label{tab:ablation}%
\end{wraptable}%

\textbf{Ablation Study.} We present the ablation study on two key design choices: the clustering method used by MajorMark and the number of message blocks $r$ in MajorMark$^+$. Table~\ref{tab:ablation} summarizes the resulting BA and PPL with message length fixed at $b=32$. For MajorMark, we evaluate two additional clustering methods beyond the default $\mathtt{KMeans}$: Agglomerative Clustering ($\mathtt{AC}$)~\citep{mullner2011modern} and Gaussian Mixture Models ($\mathtt{GMM}$)~\citep{reynolds2015gaussian}. With a strong watermarking bias ($\delta=4$), all three methods achieve comparable BA. When the bias is weaker ($\delta=2$), $\mathtt{KMeans}$ slightly outperforms the others. Overall, the performance remains stable across different clustering choices, indicating that MajorMark is robust to the selection of the clustering method. For MajorMark$^+$, we vary the number of blocks $r \in \{1, 2, 4, 8\}$, where $r=1$ treats the message as a single block. Increasing $r$ generally improves PPL, which matches our analysis of how block granularity affects the expected green list size $\gamma$. Among these settings, $r=2$ provides a good balance between BA and PPL, making it a practical choice for real-world use. Additionally, we plot the Pareto curves for different $(r,\delta)$ in Appendix~\ref{apdx: pareto_curves}.

\textbf{Additional Results and Discussions.} 
Examples of watermarked texts are provided in Appendix~\ref{apdx: examples_of_texts}. 
We discuss how our methods address practical false positive cases in Appendix~\ref{apdx: falsepostive}. 
We further discuss how extremely imbalanced messages impact our methods in Appendix~\ref{apdx: bit_imbalance}.

\section{Conclusion} \label{sec: conclusion}
We find that existing perturbation-based multi-bit watermarking methods rely on counting how frequently generated tokens fall within the reconstructed green list to decode the embedded message. This paradigm inherently constrains the size of the green list to ensure decoding accuracy. To address this, we propose MajorMark and MajorMark$^+$, both of which employ a novel majority bit-aware encoding strategy. This design enables the construction of larger green lists, as we theoretically prove. Extensive experiments demonstrate that MajorMark and MajorMark$^+$ consistently outperform existing baselines in both text utility and decoding accuracy. 

While our majority bit-aware encoding automatically determines the green list size for each message, this design imposes constraints on the flexibility of $\gamma$ in certain scenarios. In the future, it would be desirable to develop methods that preserve the advantages of MajorMark and MajorMark$^+$, namely, their independence from green list token frequencies, while allowing for a more flexible specification of $\gamma$. In addition, MajorMark$^+$ requires additional computation time to decode messages from the generated text. Although it remains more efficient than several existing approaches, further improving its decoding efficiency remains a promising direction for future work. Co-designing  the green list ratio $\gamma$ and the watermarking bias $\delta$ is also a promising future direction.

\section*{Acknowledgements and Disclosures}
This material is based upon work co-supported by the U.S. Department of Energy, Office of Science, Office of Advanced Scientific Computing Research under Contract No. DE-AC05-00OR22725. This manuscript has been co-authored by UT-Battelle, LLC under Contract No. DE-AC05-00OR22725 with the U.S. Department of Energy. The United States Government retains and the publisher, by accepting the article for publication, acknowledges that the United States Government retains a non-exclusive, paid-up, irrevocable, world-wide license to publish or reproduce the published form of this manuscript, or allow others to do so, for United States Government purposes. The Department of Energy will provide public access to these results of federally sponsored research in accordance with the DOE Public Access Plan (http://energy.gov/downloads/doe-public-access-plan).

{
    \small
    \bibliographystyle{plain}
    \bibliography{arxiv/ref}
}


\appendix
\clearpage

\section{Notation Table}
\label{sec:notation_table}
\begin{table}[t]
    \centering
    \caption{Notation table.}
    \label{tab:notation}
    \renewcommand{\arraystretch}{1.15}
    \scalebox{0.87}{
    \begin{tabular}{ll}
        \toprule
        \textbf{Symbol} & \textbf{Description} \\
        \midrule
        $\gamma$ & The green list ratio \\
        $\mathbf{m}$ & The binary multi-bit message \\
        $m_i$ & The $i$-th bit of the message $\mathbf{m}$ \\
        $b$ & The length of message $\mathbf{m}$ \\
        $f$ & The large language model \\
        $\mathbf{x}_p$ & The input prompt \\
        $\mathbf{x}$ & The unverified text \\
        $x_t$ & The $t$-th token generated by $f$ \\
        $\mathcal{V}$ & The vocabulary of $f$ \\
        $T$ & The total generation step of $f$ and the length of $\mathbf{x}_g$ \\
        $\mathbf{x}_g$ & The generated text of $f$ without watermark \\
        $\mathbf{x}_g^\prime$ & The generated text of $f$ with watermark \\
        $\mathcal{G}$ & The green list \\
        $\mathcal{R}$ & The red list \\
        $\delta$ & The watermarking bias \\
        $\mathtt{Enc} (\cdot)$ & The encoding function for a watermarking method \\
        $\mathtt{Dec} (\cdot)$ & The decoding function for a watermarking method \\
        $\mathbf{m}^\prime$ & The decoded message via $\mathtt{Dec} (\cdot)$ \\
        $\mathtt{Quality}(\cdot)$ & The quality function \\
        $\mathtt{PPL}(\cdot)$ & The conditional perplexity \\
        $\tau$ & The tolerance margin on text quality for $\mathtt{Enc} (\cdot)$ \\
        $\lambda$ & The majority bit of a message $\mathbf{m}$ \\
        $k$ & The secret hash key \\
        $\mathtt{Hash}(\cdot)$ & The hash function \\
        $s$ & The pseudo-random seed generated by $\mathtt{Hash}(\cdot)$ \\
        $\mathcal{C}$ & The cluster with larger mean produced by MajorMark \\
        $r$ & The number of blocks in MajorMark$^+$ \\
        $h_0$ or $h_1$ & The occurrences of bit $0$ or $1$ in a message \\
        $d$ & The number of bits in a message block \\
        \bottomrule
    \end{tabular}
    }
\end{table}
We present the detailed notation table in Table~\ref{tab:notation}.

\section{Additional Results}

\label{apdx: additional_results}
\label{app:full_main_table}

\textbf{Results with Uncertainty Estimates.}
We report the standard deviation and 95\% confidence interval (CI) for the results with message lengths ($b=32$) and ($b=64$) in Table~\ref{tab:main_result_table_full}. These statistics provide additional information about the variability of BA and PPL across different messages.

\begin{table*}[t]
  \centering
  \caption{
Part of results from Table~\ref{tab:main_result_table} with standard deviations and 95\% confidence intervals. The first line in each cell reports mean$_{\pm\mathrm{STD}}$, and the second line reports the 95\% confidence interval.
}
  \scalebox{0.75}{
    \begin{tabular}{l|cc|cc|cc}
    \toprule
    \footnotesize
    \multirow{2}[2]{*}{\textbf{Method}} 
    & \multicolumn{2}{c|}{$\delta=2$}   
    & \multicolumn{2}{c|}{$\delta=4$}   
    & \multicolumn{2}{c}{$\delta=6$} \\
    \cmidrule(r){2-3}\cmidrule(r){4-5}\cmidrule(r){6-7}
    & \textbf{BA}$\uparrow$ & \textbf{PPL}$\downarrow$
    & \textbf{BA}$\uparrow$ & \textbf{PPL}$\downarrow$
    & \textbf{BA}$\uparrow$ & \textbf{PPL}$\downarrow$ \\
    \midrule

    \multicolumn{7}{l}{\textit{Message length $b=32$}} \\

    MPAC
    & \makecell{$89.22_{\pm 6.66}$\\{\scriptsize [86.10, 92.34]}}
    & \makecell{$5.03_{\pm 0.70}$\\{\scriptsize [4.70, 5.36]}}
    & \makecell{$98.75_{\pm 1.82}$\\{\scriptsize [97.90, 99.60]}}
    & \makecell{$9.37_{\pm 1.47}$\\{\scriptsize [8.68, 10.06]}}
    & \makecell{$\mathbf{100.00}_{\pm 0.00}$\\{\scriptsize [100.00, 100.00]}}
    & \makecell{$15.32_{\pm 2.46}$\\{\scriptsize [14.17, 16.47]}} \\

    RSBH
    & \makecell{$89.38_{\pm 11.67}$\\{\scriptsize [83.92, 94.84]}}
    & \makecell{$4.68_{\pm 0.47}$\\{\scriptsize [4.46, 4.90]}}
    & \makecell{$97.66_{\pm 10.22}$\\{\scriptsize [92.88, 102.44]}}
    & \makecell{$6.43_{\pm 1.16}$\\{\scriptsize [5.89, 6.97]}}
    & \makecell{$97.50_{\pm 10.90}$\\{\scriptsize [92.40, 102.60]}}
    & \makecell{$8.47_{\pm 1.25}$\\{\scriptsize [7.88, 9.06]}} \\

    \rowcolor[rgb]{ .9, .9, .9}MajorMark
    & \makecell{$96.74_{\pm 3.07}$\\{\scriptsize [95.30, 98.18]}}
    & \makecell{$\mathbf{4.43}_{\pm 0.52}$\\{\scriptsize [4.19, 4.67]}}
    & \makecell{$99.06_{\pm 1.74}$\\{\scriptsize [98.25, 99.87]}}
    & \makecell{$6.32_{\pm 0.92}$\\{\scriptsize [5.89, 6.75]}}
    & \makecell{$\mathbf{100.00}_{\pm 0.00}$\\{\scriptsize [100.00, 100.00]}}
    & \makecell{$8.36_{\pm 1.35}$\\{\scriptsize [7.73, 8.99]}} \\

    \rowcolor[rgb]{ .9, .9, .9}MajorMark$^+$
    & \makecell{$\mathbf{97.81}_{\pm 2.98}$\\{\scriptsize [96.42, 99.20]}}
    & \makecell{$4.49_{\pm 0.75}$\\{\scriptsize [4.14, 4.84]}}
    & \makecell{$\mathbf{100.00}_{\pm 0.00}$\\{\scriptsize [100.00, 100.00]}}
    & \makecell{$\mathbf{6.05}_{\pm 0.84}$\\{\scriptsize [5.66, 6.44]}}
    & \makecell{$\mathbf{100.00}_{\pm 0.00}$\\{\scriptsize [100.00, 100.00]}}
    & \makecell{$\mathbf{7.90}_{\pm 1.41}$\\{\scriptsize [7.24, 8.56]}} \\

    \midrule

    \multicolumn{7}{l}{\textit{Message length $b=64$}} \\

    MPAC
    & \makecell{$81.48_{\pm 5.89}$\\{\scriptsize [78.72, 84.24]}}
    & \makecell{$4.99_{\pm 0.66}$\\{\scriptsize [4.68, 5.30]}}
    & \makecell{$93.59_{\pm 2.36}$\\{\scriptsize [92.49, 94.69]}}
    & \makecell{$8.64_{\pm 0.89}$\\{\scriptsize [8.22, 9.06]}}
    & \makecell{$96.48_{\pm 2.46}$\\{\scriptsize [95.33, 97.63]}}
    & \makecell{$15.95_{\pm 2.21}$\\{\scriptsize [14.92, 16.98]}} \\

    RSBH
    & \makecell{$81.25_{\pm 7.33}$\\{\scriptsize [77.82, 84.68]}}
    & \makecell{$4.93_{\pm 0.80}$\\{\scriptsize [4.56, 5.30]}}
    & \makecell{$90.39_{\pm 6.73}$\\{\scriptsize [87.24, 93.54]}}
    & \makecell{$6.41_{\pm 0.79}$\\{\scriptsize [6.04, 6.78]}}
    & \makecell{$97.58_{\pm 4.24}$\\{\scriptsize [95.60, 99.56]}}
    & \makecell{$9.22_{\pm 1.01}$\\{\scriptsize [8.75, 9.69]}} \\

    \rowcolor[rgb]{ .9, .9, .9}MajorMark
    & \makecell{$83.59_{\pm 5.94}$\\{\scriptsize [80.81, 86.37]}}
    & \makecell{$4.78_{\pm 0.80}$\\{\scriptsize [4.41, 5.15]}}
    & \makecell{$93.83_{\pm 4.35}$\\{\scriptsize [91.79, 95.87]}}
    & \makecell{$\mathbf{6.11}_{\pm 1.17}$\\{\scriptsize [5.56, 6.66]}}
    & \makecell{$98.67_{\pm 1.24}$\\{\scriptsize [98.09, 99.25]}}
    & \makecell{$8.82_{\pm 1.36}$\\{\scriptsize [8.18, 9.46]}} \\

    \rowcolor[rgb]{ .9, .9, .9}MajorMark$^+$
    & \makecell{$\mathbf{86.95}_{\pm 7.64}$\\{\scriptsize [83.37, 90.53]}}
    & \makecell{$\mathbf{4.74}_{\pm 0.86}$\\{\scriptsize [4.34, 5.14]}}
    & \makecell{$\mathbf{97.81}_{\pm 2.44}$\\{\scriptsize [96.67, 98.95]}}
    & \makecell{$6.21_{\pm 0.82}$\\{\scriptsize [5.83, 6.59]}}
    & \makecell{$\mathbf{99.69}_{\pm 1.36}$\\{\scriptsize [99.05, 100.33]}}
    & \makecell{$\mathbf{7.93}_{\pm 1.27}$\\{\scriptsize [7.34, 8.52]}} \\

    \bottomrule
    \end{tabular}
  }
  \label{tab:main_result_table_full}
\end{table*}

\begin{table*}[t]
  \centering
    \caption{Performance of MajorMark and MajorMark$^+$ on Qwen$2.5$-$7$B, Gemma-$2$B, and LLaMA-$3.1$-$8$B with $b \in \{32, 64\}$ and $\delta \in \{2, 4\}$. The PPL for non-watermarked texts generated by Qwen$2.5$-$7$B, Gemma-$2$B, and LLaMA-$3.1$-$8$B are $4.97$, $5.71$, and $4.27$, respectively.}
  \scalebox{0.73}{
    \begin{tabular}{l|l|ccc|ccc|ccc}
    \toprule
    \footnotesize
     \multirow{2}[2]{*}{\textbf{Model}} & \multirow{2}[2]{*}{\textbf{Method}} & \multicolumn{3}{c|}{$\delta=2$} & \multicolumn{3}{c|}{$\delta=4$} & \multirow{2}[2]{*}{\textbf{\makecell*[c]{Avg. \\ BA$\uparrow$}}} & \multirow{2}[2]{*}{\textbf{\makecell*[c]{Avg. \\ PPL$\downarrow$}}} & \multirow{2}[2]{*}{\textbf{\makecell*[c]{Avg. \\ Top-$\mathbf{5}$$\uparrow$}}} \\
    \cmidrule(r){3-5}\cmidrule(r){6-8}
            &    & \textbf{BA}$\uparrow$    & \textbf{PPL}$\downarrow$   & \textbf{Top-$\mathbf{5}$}$\uparrow$ & \textbf{BA}$\uparrow$    & \textbf{PPL}$\downarrow$   & \textbf{Top-$\mathbf{5}$}$\uparrow$ \\
    \midrule
    \multicolumn{8}{l}{\textit{Message length $b=32$}} \\
    \multirow{2}[2]{*}{Qwen$2.5$-$7$B} 
& MajorMark & $96.25$ & $6.69$ & $86.99$ & $99.22$ & $8.49$ & $85.29$ & $97.74$ & $7.59$ & $86.14$ \\
& MajorMark$^+$ & $98.75$ & $6.31$ & $88.39$ & $100.00$ & $8.85$ & $84.77$ & $\mathbf{99.38}$ & $\mathbf{7.58}$ & $\mathbf{86.58}$ \\

\cmidrule(r){1-11}

\multirow{2}[2]{*}{Gemma-$2$B} & MajorMark & $96.41$ & $7.02$ & $87.16$ & $99.06$ & $9.80$ & $83.38$ & $97.74$ & $8.41$ & $\mathbf{85.27}$ \\
& MajorMark$^+$ & $99.38$ & $6.95$ & $86.02$ & $100.00$ & $9.69$ & $84.36$ & $\mathbf{99.69}$ & $\mathbf{8.32}$ & $85.19$ \\

\cmidrule(r){1-11}

\multirow{2}[2]{*}{LLaMA-$3.1$-$8$B} & MajorMark & $94.84$ & $5.69$ & $90.85$ & $99.69$ & $7.94$ & $88.56$ & $97.27$ & $6.82$ & $89.71$ \\
& MajorMark$^+$ & $97.81$ & $5.78$ & $90.90$ & $100.00$ & $7.57$ & $88.75$ & $\mathbf{98.91}$ & $\mathbf{6.68}$ & $\mathbf{89.8}3$ \\

\midrule \midrule

\multicolumn{8}{l}{\textit{Message length $b=64$}} \\ 
\multirow{2}[2]{*}{Qwen$2.5$-$7$B} 
& MajorMark &  $88.83$ & $6.43$ & $88.02$ & $96.72$ & $9.19$ & $84.27$ & $92.78$ & $7.81$ & $86.14$ \\
& MajorMark$^+$ & $92.03$ & $6.31$ & $88.20$ & $99.22$ & $9.00$ & $84.25$ & $\mathbf{95.63}$ & $\mathbf{7.66}$ & $\mathbf{86.23}$ \\

\cmidrule(r){1-11}

\multirow{2}[2]{*}{Gemma-$2$B} & MajorMark & $88.44$ & $7.54$ & $85.82$ & $96.72$ & $10.79$ & $82.72$ & $92.58$ & $9.17$ & $84.27$ \\
& MajorMark$^+$ & $93.75$ & $7.41$ & $85.93$ & $99.38$ & $9.82$ & $82.75$ & $\mathbf{96.56}$ & $\mathbf{8.62}$ & $\mathbf{84.34}$ \\

\cmidrule(r){1-11}

\multirow{2}[2]{*}{LLaMA-$3.1$-$8$B} & MajorMark & $86.17$ & $5.33$ & $92.11$ & $95.94$ & $7.58$ & $88.48$ & $91.06$ & $\mathbf{6.46}$ & $\mathbf{90.30}$ \\
& MajorMark$^+$ & $92.50$ & $5.72$ & $91.32$ & $99.06$ & $7.67$ & $88.51$ & $\mathbf{95.78}$ & $6.70$ & $89.92$ \\

    \bottomrule
    \end{tabular}%
    }
  \label{tab:qwen}%
\end{table*}

\textbf{Results on Additional LLMs}
We additionally report the performance of our methods on three other popular language models: Qwen$2.5$-$7$B~\citep{qwen2}, Gemma-$2$B~\citep{team2024gemma}, and LLaMA-$3.1$-$8$B~\citep{grattafiori2024llama}. Specifically, we present the BA, PPL, and Top-5 hit rate results under message lengths $b \in \{32, 64\}$ and watermarking biases $\delta \in \{2, 4\}$ in Table~\ref{tab:qwen}. Note that for evaluating PPL, we use the larger models Qwen$2.5$-$32$B, Gemma-$7$B, and LLaMA-$3.1$-$70$B. As shown, both MajorMark and MajorMark$^+$ perform well across the board. 
These results demonstrate the strong generalization ability of MajorMark and MajorMark$^+$ across different LLM architectures and capacities.

\textbf{Additional Results on OpenGen and Essays Datasets} 
We conduct a comprehensive evaluation of MPAC, RSBH, and our two proposed methods, MajorMark and MajorMark$^+$, on the OpenGen~\citep{opengen} and Essays~\citep{essays} datasets. Results for message lengths $b \in \{32, 64\}$ and watermarking biases $\delta \in \{2, 4, 6\}$ are reported in Table~\ref{tab:opgen} and Table~\ref{tab:essay}, respectively. 

Consistent with the trends observed on the C$4$ dataset, our methods consistently outperform the state-of-the-art baselines. In particular, MajorMark$^+$ achieves the highest average BA and Top-5 hit rate, while maintaining the lowest PPL across all settings. This demonstrates its strong ability to preserve the utility of watermarked text while ensuring high decoding accuracy. For example, on the Essays dataset with $b = 64$, MajorMark$^+$ achieves a BA of $95.21\%$, which is $+5.99\%$ and $+5.52\%$ higher than MPAC and RSBH, respectively. In terms of PPL, MajorMark$^+$ obtains the lowest score of $6.03$, outperforming MPAC and RSBH by margins of $3.20$ and $0.19$, respectively. These results further confirm the strong generalization ability of both MajorMark and MajorMark$^+$ across diverse datasets, highlighting the practical effectiveness of our methods in varying domains.

\begin{table*}[t]
  \centering
  \caption{
Performance comparison of watermarking methods (MPAC, RSBH, MajorMark, and MajorMark$^+$) on the OpenGen dataset with message lengths $b \in \{32, 64\}$ and watermarking biases $\delta \in \{2, 4, 6\}$.
}
  \scalebox{0.72}{
    \begin{tabular}{l|ccc|ccc|ccc|ccc}
    \toprule
    \footnotesize
     \multirow{2}[2]{*}{\textbf{Method}} & \multicolumn{3}{c|}{$\delta=2$}   & \multicolumn{3}{c|}{$\delta=4$}   & \multicolumn{3}{c|}{$\delta=6$} & \multirow{2}[2]{*}{\textbf{\makecell*[c]{Avg. \\ BA$\uparrow$}}} & \multirow{2}[2]{*}{\textbf{\makecell*[c]{Avg. \\ PPL$\downarrow$}}} & \multirow{2}[2]{*}{\textbf{\makecell*[c]{Avg. \\ Top-$\mathbf{5}$$\uparrow$}}} \\
    \cmidrule(r){2-4}\cmidrule(r){5-7}\cmidrule(r){8-10}
               & \textbf{BA}$\uparrow$    & \textbf{PPL}$\downarrow$   & \textbf{Top-$\mathbf{5}$}$\uparrow$ & \textbf{BA}$\uparrow$    & \textbf{PPL}$\downarrow$   & \textbf{Top-$\mathbf{5}$}$\uparrow$ & \textbf{BA}$\uparrow$    & \textbf{PPL}$\downarrow$   & \textbf{Top-$\mathbf{5}$}$\uparrow$ \\
    \midrule
    \multicolumn{8}{l}{\textit{Message length $b=32$}} \\
         MPAC  & $86.56$ & $4.77$ & $91.26$ & $97.34$ & $9.23$ & $81.86$ & $99.38$ & $15.08$ & $76.41$ & $94.43$ & $9.69$ & $83.18$ \\
         RSBH  & $84.53$ & $4.68$ & $92.47$ & $96.56$ & $6.43$ & $89.98$ & $97.66$ & $8.64$ & $88.47$ & $92.92$ & $6.58$ & $90.31$ \\
         \rowcolor[rgb]{  .9, .9, .9}MajorMark 
              & $90.94$ & $4.35$ & $92.83$ & $97.81$ & $6.02$ & $90.01$ & $99.38$ & $8.51$ & $87.95$ & $96.04$ & $6.29$ & $90.26$\\
         \rowcolor[rgb]{  .9, .9, .9}MajorMark$^+$ 
              & $\mathbf{92.03}$ & $\mathbf{4.09}$ & $\mathbf{93.85}$ & $\mathbf{99.38}$ & $\mathbf{5.63}$ & $\mathbf{91.16}$ & $\mathbf{100.00}$ & $\mathbf{7.96}$ & $\mathbf{88.90}$ & $\mathbf{97.14}$ & $\mathbf{5.89}$ & $\mathbf{91.30}$ \\
    \midrule \midrule
    \multicolumn{8}{l}{\textit{Message length $b=64$}} \\
         MPAC  & $78.12$ & $4.83$ & $90.64$ & $90.70$ & $8.95$ & $82.52$ & $96.33$ & $15.79$ & $75.50$ & $88.38$ & $9.86$ & $82.89$ \\
         RSBH  & $77.11$ & $4.53$ & $\mathbf{93.05}$ & $89.06$ & $6.33$ & $\mathbf{90.25}$ & $96.95$ & $8.59$ & $87.79$ & $87.71$ & $6.48$ & $90.36$ \\
         \rowcolor[rgb]{  .9, .9, .9}MajorMark   
              & $\mathbf{84.06}$ & $4.77$ & $92.24$ & $93.44$ & $\mathbf{6.14}$ & $89.97$ & $96.02$ & $8.42$ & $88.22$ & $91.17$ & $6.44$ & $90.14$\\
         \rowcolor[rgb]{  .9, .9, .9}MajorMark$^+$   
              & $79.92$ & $\mathbf{4.41}$ & $92.62$ & $\mathbf{97.66}$ & $6.37$ & $89.73$ & $\mathbf{99.06}$ & $\mathbf{8.01}$ & $\mathbf{88.98}$ & $\mathbf{92.21}$ & $\mathbf{6.26}$ & $\mathbf{90.44}$ \\
    \bottomrule
    \end{tabular}%
    }
  \label{tab:opgen}%
\end{table*}

\begin{table*}[t]
  \centering
  \caption{
Performance comparison of watermarking methods (MPAC, RSBH, MajorMark, and MajorMark$^+$) on the Essays dataset with message lengths $b \in \{32, 64\}$ and watermarking biases $\delta \in \{2, 4, 6\}$.
}
  \scalebox{0.68}{
    \begin{tabular}{l|ccc|ccc|ccc|ccc}
    \toprule
    \footnotesize
     \multirow{2}[2]{*}{\textbf{Method}} & \multicolumn{3}{c|}{$\delta=2$}   & \multicolumn{3}{c|}{$\delta=4$}   & \multicolumn{3}{c|}{$\delta=6$} & \multirow{2}[2]{*}{\textbf{\makecell*[c]{Avg. \\ BA$\uparrow$}}} & \multirow{2}[2]{*}{\textbf{\makecell*[c]{Avg. \\ PPL$\downarrow$}}} & \multirow{2}[2]{*}{\textbf{\makecell*[c]{Avg. \\ Top-$\mathbf{5}$$\uparrow$}}} \\
    \cmidrule(r){2-4}\cmidrule(r){5-7}\cmidrule(r){8-10}
              & \textbf{BA}$\uparrow$    & \textbf{PPL}$\downarrow$   & \textbf{Top-$\mathbf{5}$}$\uparrow$ & \textbf{BA}$\uparrow$    & \textbf{PPL}$\downarrow$   & \textbf{Top-$\mathbf{5}$}$\uparrow$ & \textbf{BA}$\uparrow$    & \textbf{PPL}$\downarrow$   & \textbf{Top-$\mathbf{5}$}$\uparrow$ \\
    \midrule
    \multicolumn{8}{l}{\textit{Message length $b=32$}} \\
         MPAC  & $88.44$ & $4.29$ & $92.18$ & $98.75$ & $9.28$ & $80.54$ & $99.69$ & $14.91$ & $75.30$ & $95.63$ & $9.49$ & $82.67$ \\
         RSBH  & $89.22$ & $4.23$ & $93.31$ & $97.66$ & $6.33$ & $88.98$ & $97.66$ & $7.78$ & $87.78$ & $94.85$ & $6.11$ & $90.02$ \\
         \rowcolor[rgb]{  .9, .9, .9}MajorMark 
              & $91.41$ & $\mathbf{4.12}$ & $93.24$ & $98.28$ & $5.98$ & $89.60$ & $99.53$ & $7.71$ & $87.70$ & $96.41$ & $5.94$ & $90.18$ \\
         \rowcolor[rgb]{  .9, .9, .9}MajorMark$^+$ 
              & $\mathbf{96.72}$ & $4.21$ & $\mathbf{93.37}$ & $\mathbf{100.00}$ & $\mathbf{5.83}$ & $\mathbf{90.44}$ & $\mathbf{100.00}$ & $\mathbf{7.40}$ & $\mathbf{88.92}$ & $\mathbf{98.91}$ & $\mathbf{5.81}$ & $\mathbf{90.91}$ \\
    \midrule \midrule
    \multicolumn{8}{l}{\textit{Message length $b=64$}}  \\
         MPAC  & $78.36$ & $4.33$ & $92.85$ & $92.58$ & $9.12$ & $80.94$ & $96.72$ & $14.23$ & $75.83$ & $89.22$ & $9.23$ & $83.21$ \\
         RSBH  & $76.72$ & $\mathbf{4.05}$ & $93.50$ & $93.67$ & $6.11$ & $89.25$ & $98.67$ & $8.51$ & $87.16$ & $89.69$ & $6.22$ & $89.97$ \\
         \rowcolor[rgb]{  .9, .9, .9}MajorMark   
              & $82.73$ & $4.33$ & $92.56$ & $94.84$ & $6.07$ & $89.48$ & $96.88$ & $8.46$ & $87.07$ & $91.48$ & $6.29$ & $89.70$\\
         \rowcolor[rgb]{  .9, .9, .9}MajorMark$^+$   
              & $\mathbf{87.19}$ & $4.14$ & $\mathbf{93.56}$ & $\mathbf{98.75}$ & $\mathbf{5.99}$ & $\mathbf{89.83}$ & $\mathbf{99.69}$ & $\mathbf{7.95}$ & $\mathbf{87.42}$ & $\mathbf{95.21}$ & $\mathbf{6.03}$ & $\mathbf{90.27}$ \\
    \bottomrule
    \end{tabular}%
    }
  \label{tab:essay}%
\end{table*}

\begin{table*}[t]
  \centering
  \caption{
Performance comparison of watermarking methods on the WritingPrompts dataset for the story generation task with message lengths $b \in \{32, 64\}$ and watermarking biases $\delta \in \{4, 6\}$.
}
  \scalebox{0.75}{
    \begin{tabular}{l|ccc|ccc|ccc}
    \toprule
    \footnotesize
     \multirow{2}[2]{*}{\textbf{Method}}  
      & \multicolumn{3}{c|}{$\delta=4$}   
      & \multicolumn{3}{c|}{$\delta=6$} 
      & \multirow{2}[2]{*}{\textbf{\makecell*[c]{Avg. \\ BA$\uparrow$}}} 
      & \multirow{2}[2]{*}{\textbf{\makecell*[c]{Avg. \\ PPL$\downarrow$}}} 
      & \multirow{2}[2]{*}{\textbf{\makecell*[c]{Avg. \\ Top-$\mathbf{5}$$\uparrow$}}} \\
    \cmidrule(r){2-4}\cmidrule(r){5-7}
               & \textbf{BA}$\uparrow$    
                  & \textbf{PPL}$\downarrow$   
                  & \textbf{Top-$\mathbf{5}$}$\uparrow$ 
                  & \textbf{BA}$\uparrow$    
                  & \textbf{PPL}$\downarrow$   
                  & \textbf{Top-$\mathbf{5}$}$\uparrow$ \\
    \midrule
    \multicolumn{8}{l}{\textit{Message length $b=32$}} \\
         MPAC  & $84.69$ & $4.69$ & $96.09$ & $98.28$ & $13.65$ & $85.41$ & $91.49$ & $9.17$ & $90.75$ \\
         RSBH  & $83.28$ & $4.17$ & $98.17$ & $96.41$ & $6.20$ & $96.81$ & $89.85$ & $5.19$ & $97.49$ \\
         \rowcolor[rgb]{  .9, .9, .9}MajorMark & $89.38$ & $4.09$ & $98.70$ & $96.41$ & $6.19$ & $97.12$ & $92.90$ & $5.14$ & $97.91$ \\
         \rowcolor[rgb]{  .9, .9, .9}MajorMark$^+$ & $\mathbf{91.88}$ & $\mathbf{3.91}$ & $\mathbf{98.72}$ & $\mathbf{98.44}$ & $\mathbf{5.52}$ & $\mathbf{97.51}$ & $\mathbf{95.16}$ & $\mathbf{4.72}$ & $\mathbf{98.12}$ \\
    \midrule \midrule
    \multicolumn{8}{l}{\textit{Message length $b=64$}} \\
         MPAC  & $76.33$ & $4.65$ & $96.02$ & $91.33$ & $11.32$ & $87.04$ & $83.83$ & $7.99$ & $91.53$ \\
         RSBH  & $76.41$ & $4.14$ & $98.36$ & $85.39$ & $6.18$ & $97.06$ & $80.90$ & $5.16$ & $97.71$ \\
         \rowcolor[rgb]{  .9, .9, .9}MajorMark & $80.94$ & $4.05$ & $\mathbf{98.93}$ & $93.05$ & $6.08$ & $97.13$ & $87.00$ & $5.07$ & $98.03$ \\
         \rowcolor[rgb]{  .9, .9, .9}MajorMark$^+$ & $\mathbf{83.83}$ & $\mathbf{4.00}$ & $98.69$ & $\mathbf{94.53}$ & $\mathbf{5.63}$ & $\mathbf{97.40}$ & $\mathbf{89.18}$ & $\mathbf{4.82}$ & $\mathbf{98.05}$ \\
    \bottomrule
    \end{tabular}%
    }
  \label{tab: story_generation}%
\end{table*}

\textbf{Results on Story Generation Task.} 
In this section, we evaluate existing methods alongside our two proposed approaches on the story generation task. Specifically, we use the WritingPrompts dataset~\citep{writingprompt} with the LLaMA-$2$-$7$B-chat model~\citep{llama2}. The chat prompt is designed to encourage coherent and imaginative story generation and is defined as follows:
\begin{quote}
\texttt{[System]} \textit{You are a helpful assistant that writes engaging and coherent stories.}

\texttt{[User]} \textit{Please write a detailed and imaginative short story based on the following prompt: [prompt]}
\end{quote}

Results for message lengths $b \in \{32, 64\}$ and watermarking biases $\delta \in \{4, 6\}$ are summarized in Table~\ref{tab: story_generation}. Our methods, MajorMark and MajorMark$^+$, consistently achieve strong performance on the story generation task. In particular, across both watermarking biases, MajorMark$^+$ yields the highest average BA and Top-$5$ hit ratios, as well as the lowest PPL, indicating its effectiveness in embedding reliable watermarks while preserving text quality. Although MajorMark performs slightly worse than MajorMark$^+$, it still outperforms existing state-of-the-art baselines, demonstrating its robustness and practicality.

\begin{table*}[t]
  \centering
  \caption{
Performance comparison of watermarking methods on the CNN/DailyMail dataset for the text summarization task with message lengths $b \in \{32, 64\}$ and watermarking biases $\delta \in \{4, 6\}$.
}
  \scalebox{0.75}{
    \begin{tabular}{l|ccc|ccc|ccc}
    \toprule
    \footnotesize
     \multirow{2}[2]{*}{\textbf{Method}}  
      & \multicolumn{3}{c|}{$\delta=4$}   
      & \multicolumn{3}{c|}{$\delta=6$} 
      & \multirow{2}[2]{*}{\textbf{\makecell*[c]{Avg. \\ BA$\uparrow$}}} 
      & \multirow{2}[2]{*}{\textbf{\makecell*[c]{Avg. \\ PPL$\downarrow$}}} 
      & \multirow{2}[2]{*}{\textbf{\makecell*[c]{Avg. \\ Top-$\mathbf{5}$$\uparrow$}}} \\
    \cmidrule(r){2-4}\cmidrule(r){5-7}
               & \textbf{BA}$\uparrow$    
                  & \textbf{PPL}$\downarrow$   
                  & \textbf{Top-$\mathbf{5}$}$\uparrow$ 
                  & \textbf{BA}$\uparrow$    
                  & \textbf{PPL}$\downarrow$   
                  & \textbf{Top-$\mathbf{5}$}$\uparrow$ \\
    \midrule
    \multicolumn{8}{l}{\textit{Message length $b=32$}} \\
         MPAC  & $83.12$ & $6.36$ & $96.68$ & $96.41$ & $12.25$ & $90.48$ & $89.77$ & $9.31$ & $93.58$ \\
         RSBH  & $85.00$ & $5.81$ & $98.83$ & $\mathbf{97.34}$ & $8.07$ & $97.24$ & $91.17$ & $6.94$ & $98.04$ \\
         \rowcolor[rgb]{  .9, .9, .9}MajorMark & $88.12$ & $5.72$ & $99.02$ & $95.78$ & $7.70$ & $98.00$ & $91.95$ & $6.71$ & $98.51$ \\
         \rowcolor[rgb]{  .9, .9, .9}MajorMark$^+$ & $\mathbf{89.53}$ & $\mathbf{5.72}$ & $\mathbf{99.10}$ & $95.16$ & $\mathbf{7.13}$ & $\mathbf{98.52}$ & $\mathbf{92.35}$ & $\mathbf{6.43}$ & $\mathbf{98.81}$ \\
    \midrule \midrule
    \multicolumn{8}{l}{\textit{Message length $b=64$}} \\
         MPAC  & $78.67$ & $6.55$ & $97.06$ & $91.33$ & $12.86$ & $90.09$ & $85.00$ & $9.71$ & $93.58$ \\
         RSBH  & $73.05$ & $5.99$ & $98.81$ & $84.92$ & $7.87$ & $97.31$ & $78.99$ & $6.93$ & $98.06$ \\
         \rowcolor[rgb]{  .9, .9, .9}MajorMark & $83.36$ & $5.73$ & $99.02$ & $89.30$ & $8.18$ & $97.33$ & $86.33$ & $6.95$ & $98.18$ \\
         \rowcolor[rgb]{  .9, .9, .9}MajorMark$^+$ & $\mathbf{84.22}$ & $\mathbf{5.51}$ & $\mathbf{99.39}$ & $\mathbf{91.72}$ & $\mathbf{7.33}$ & $\mathbf{98.01}$ & $\mathbf{87.97}$ & $\mathbf{6.42}$ & $\mathbf{98.70}$ \\
    \bottomrule
    \end{tabular}%
    }
  \label{tab: text_summarize}%
\end{table*}

\textbf{Results on Text Summarization Task.} 
In this section, we evaluate existing methods alongside our two proposed approaches on the text summarization task. Specifically, we use the CNN/DailyMail dataset~\citep{cnndaily} with the LLaMA-$2$-$7$B-chat model as the backbone. The chat prompt is designed to instruct the model to generate concise and faithful summaries given an input article, using the following format:
\begin{quote}
\texttt{[System]} \textit{You are a helpful assistant specialized in summarization. You take a document and write a concise, faithful summary.}

\texttt{[User]} \textit{Please summarize the following article in a few sentences: [article]}
\end{quote}

Results for message lengths $b \in \{32, 64\}$ and watermarking biases $\delta \in \{4, 6\}$ are summarized in Table~\ref{tab: text_summarize}. Similar to the results observed for text completion and story generation, our methods generally achieve both the highest text quality and the highest decoding accuracy across both watermarking bias settings. These findings demonstrate the strong generalization ability of our proposed methods, MajorMark and MajorMark$^+$, across diverse natural language processing tasks. This highlights their potential as practical candidates for real-world watermarking deployment.

\setlength{\intextsep}{0mm}%
\begin{wraptable}{r}{0.51\textwidth}
    \centering
    \caption{
    Statistical significance test over $100$ users on Qwen$3$-$8$B. We report the mean and standard deviation of each metric across users.
    }
    \label{tab:statistical_significance}
    \resizebox{\linewidth}{!}{
    \begin{tabular}{l|cc|cc}
        \toprule
        \textbf{Metric} 
        & \textbf{MajorMark$^+$} 
        & \textbf{BiMark} 
        & \textbf{$W$ Statistic} 
        & \textbf{$p$-value} \\
        \midrule
        PPL & $4.75 \pm 0.95$ & $5.05 \pm 1.23$ & $1633.0$ & $0.00108$ \\
        BA  & $98.50 \pm 4.41$ & $96.66 \pm 4.66$ & $1094.0$ & $0.00002$ \\
        \bottomrule
    \end{tabular}
    }
\end{wraptable}
We further evaluate whether the observed improvements are statistically significant under the $100$-user setting. Specifically, we conduct a one-sided Wilcoxon signed-rank test on the paired per-user PPL and BA results, comparing MajorMark$^+$ against BiMark on Qwen$3$-$8$B. For PPL, the alternative hypothesis is that MajorMark$^+$ achieves lower PPL than BiMark. For BA, the alternative hypothesis is that MajorMark$^+$ achieves higher BA than BiMark. The results are summarized in Table~\ref{tab:statistical_significance}. MajorMark$^+$ achieves a statistically significant reduction in PPL, with $p=0.00108<0.05$, and a statistically significant improvement in BA, with $p=0.00002<0.05$. These results indicate that the improvements of MajorMark$^+$ over BiMark are unlikely to be caused by random variation among users.

\section{Discussion on Spoofing Attack} \label{apdx: details_of_attacks}

\textbf{Spoofing Attack.}Watermark spoofing, introduced by \citep{spoofingattack}, attempts to forge text that is falsely detected as watermarked, potentially harming the LLM provider’s reputation. However, as noted in~\citep{stealthink}, multi-bit watermarking embeds diverse message signals across tokens, which significantly increases the complexity of producing successful spoofs. Furthermore, \citep{rsbh} show that when the watermarking method employs a hash function modeled as a random oracle (ROM)~\citep{bellare1993random}, spoofing attacks become ineffective. Consequently, we do not consider spoofing attacks in this work.

\section{Security Analysis of the Majority-Bit Structure}
\label{app:majority_security}
\textbf{Information Revealed by the Green-List Ratio.}
We first consider whether the message-dependent green-list ratio $\gamma$ may reveal information about the embedded message. If $\gamma$ were directly observable, it would reveal the magnitude of the majority margin, since $|h_0-h_1| = b(2\gamma-1),$
where $h_0$ and $h_1$ denote the numbers of $0$s and $1$s in the message, respectively. However, $\gamma$ does not reveal the value of the majority bit $\lambda$. For uniformly sampled messages, the majority margin is strongly concentrated around balanced messages. Its Shannon entropy is approximately $2.69$ bits for $b=32$ and $3.15$ bits for $b=64$, indicating that even exact knowledge of $\gamma$ reveals only limited information about the original $b$-bit message.

More importantly, $\gamma$ is not directly observable from the generated text without the secret key. The token-to-shard mapping is determined by the keyed, context-dependent hash function, and therefore an adversary without the key cannot directly reconstruct the shards or identify which shards form the green list. We nevertheless note that $\gamma$ may potentially be inferred approximately from distributional characteristics of the generated text, since a larger green-list ratio generally requires less distributional perturbation.

\textbf{Key-Free Watermark Stealing.}
The regular shard structure also does not directly provide a fixed target for key-free watermark stealing. In MajorMark, the token-to-shard mapping is generated from the secret key, the preceding context, and the majority bit, and therefore changes across generation steps. As a result, shard indices are not consistently aligned from an adversary's perspective. Moreover, for randomly generated messages, no particular shard index is systematically favored across messages, since each shard has the same probability of belonging to the green list. These properties make it difficult for an attacker without the secret key to identify a fixed shard-level pattern that can be reused for watermark stealing.

\section{Semantic and Lexical Fidelity of Watermarked Texts} \label{apdx: semantic_fidelity}

\setlength{\intextsep}{0mm}%
\begin{wraptable}{r}{0.70\textwidth}
\centering
\caption{Semantic and lexical fidelity (BERTScore $\uparrow$ / ROUGE-1 $\uparrow$ / ROUGE-Lsum $\uparrow$) for different bias values $\delta$ with $b=32$.}
\label{tab:r4}
\scalebox{0.78}{
\begin{tabular}{l|c|c|c}
\toprule
\textbf{Method} & $\boldsymbol{\delta=2}$ & $\boldsymbol{\delta=4}$ & $\boldsymbol{\delta=6}$ \\
\midrule
MPAC & 
$\;0.8376 / 0.33 / 0.29\;$ &
$\;0.8210 / 0.28 / 0.25\;$ &
$\;0.8105 / 0.26 / 0.22\;$ \\
RSBH &
$\;0.8384 / 0.35 / 0.31\;$ &
$\;0.8316 / 0.33 / 0.29\;$ &
$\;0.8234 / 0.30 / 0.26\;$ \\
\rowcolor[rgb]{.9, .9, .9}MajorMark &
$\;0.8415 / 0.37 / 0.33\;$ &
$\;0.8305 / 0.33 / 0.28\;$ &
$\;0.8269 / 0.31 / 0.28\;$ \\
\rowcolor[rgb]{.9,.9,.9}MajorMark$^+$ &
$\;0.8449 / 0.36 / 0.32\;$ &
$\;0.8381 / 0.34 / 0.29\;$ &
$\;0.8311 / 0.33 / 0.29\;$ \\
\bottomrule
\end{tabular}
}
\end{wraptable}
Recall that in evaluation, we mainly use PPL as the metric for evaluating the quality of watermarked texts. To provide a more complete analysis, we further evaluate BERTScore (semantic similarity) and ROUGE-1 / ROUGE-Lsum (lexical and structural overlap) between watermarked and unwatermarked outputs generated from the same prompts. 
Table~\ref{tab:r4} reports results under varying values of $\delta$ with $b=32$. Across all settings, MajorMark$^+$ achieves the highest BERTScore, indicating stronger semantic preservation than MPAC and RSBH. 
In addition, the ROUGE metrics show that both MajorMark and MajorMark$^+$ maintain a high degree of lexical and structural similarity to the unwatermarked outputs. 
These results suggest that the large green list in our methods allows the model to retain its natural word choices and phrasing, rather than forcing substitutions that may degrade semantic quality.

\section{Examples of Generated Texts}\label{apdx: examples_of_texts}
Table~\ref{tab:generated_text} presents examples of generated text produced by different watermarking methods for a given prompt, with watermarking bias $\delta=4$ and message length $b=32$. As shown, both MajorMark and MajorMark$^+$ yield lower PPL, indicating better fluency in the generated text.

\section{False Positive Control} \label{apdx: falsepostive}

False positives are a practical concern in multi-bit watermarking: since the decoder always returns some message, even human-written text may be decoded into a seemingly valid message if no detection step is applied. We therefore equip both methods with a dispersion-based detection test that rejects unwatermarked text before message recovery, and calibrate its threshold on genuine human-written text.

For a candidate decoding configuration $c$, let $\mathbf{o}_c$ denote the shard-wise token occurrence vector reconstructed under $c$, and define its dispersion as $\sigma(\mathbf{o}_c) = \mathrm{std}(\mathbf{o}_c)$. Only the correct configuration reproduces the token-to-shard mappings used during encoding. Hence, watermarked text yields a markedly larger dispersion under the correct configuration than under incorrect ones, whereas human-written text yields similarly near-uniform occurrence vectors under all configurations.

\begin{itemize}[leftmargin=*]
    \item \textbf{MajorMark.} The candidate configurations are $\lambda' \in \{0, 1\}$. We define the detection score as $S = |\sigma(\mathbf{o}_{0}) - \sigma(\mathbf{o}_{1})|$. If $S < \eta$, the decoder declares the text unwatermarked; otherwise, it proceeds with message decoding.
    \item \textbf{MajorMark$^+$.} For each block $p$, we compute $S_p$ as the difference between the dispersion of the most dispersed configuration $(\lambda', h')$ and the average dispersion of the remaining configurations. If $\min_p S_p < \eta_+$, i.e., any block fails the test, the decoder declares the text unwatermarked; otherwise, it proceeds with block-wise message recovery.
\end{itemize}

\textbf{Calibration protocol.} We calibrate the thresholds using human-written text only. Specifically, we compute the detection scores on $4{,}000$ human-written records from the C4 dataset and set each threshold to the empirical $99$th percentile of the scores, corresponding to a target FPR of $1\%$. We use message length $b=32$, watermarking bias $\delta=2$. This yields $\eta = 2.15$ for MajorMark and $\eta_+ = 3.30$ for MajorMark$^+$. We then freeze both thresholds and evaluate the FPR on a disjoint held-out set of $1{,}000$ human-written C4 records, and the TPR, BA, and exact-message recovery rate (EMR) on the corresponding watermarked test set. 

\begin{wraptable}{r}{0.6\textwidth}
    \centering
    \caption{Detection and decoding performance at a calibrated target FPR of $1\%$ ($b=32$, $\delta=2$).}
    \label{tab:low_fpr}
    \resizebox{1\linewidth}{!}{
    \begin{tabular}{ll|cccc}
        \toprule
        \textbf{Dataset} & \textbf{Method} & \textbf{FPR}$\downarrow$ & \textbf{TPR}$\uparrow$ & \textbf{BA}$\uparrow$ & \textbf{EMR}$\uparrow$ \\
        \midrule
        \multirow{2}{*}{C4 (held-out)} 
        & MajorMark     & $0.90$ & $96.40$ & $96.87$ & $26.45$ \\
        & MajorMark$^+$ & $1.20$ & $95.80$ & $99.08$ & $71.67$ \\
        \midrule
        \multirow{2}{*}{OpenGen (transfer)} 
        & MajorMark     & $1.30$ & $94.80$ & $94.74$ & $20.89$ \\
        & MajorMark$^+$ & $0.80$ & $94.50$ & $95.95$ & $56.25$ \\
        \bottomrule
    \end{tabular}
    }
\end{wraptable}

\textbf{Results.} As shown in Table~\ref{tab:low_fpr}, the empirical FPRs of $0.90\%$ and $1.20\%$ on held-out human-written text closely match the $1\%$ target. At these thresholds, MajorMark and MajorMark$^+$ retain TPRs of $96.40\%$ and $95.80\%$, respectively, while BA remains comparable to that obtained without detection in Table~\ref{tab:main_result_table}. Thus, calibration controls false positives without degrading decoding performance. In terms of exact recovery, the fraction of watermarked texts that are both detected and attributed to the exact message is $25.50\%$ for MajorMark and $68.66\%$ for MajorMark$^+$. Since the thresholds are calibrated on human-written text only, they do not depend on $\delta$. Applying the same thresholds at $\delta=4$ and $\delta=6$ leaves the FPR unchanged and increases the TPR to $100\%$, as a larger $\delta$ increases the dispersion of watermarked text.

\textbf{Transfer across corpora.} To test whether the calibrated thresholds transfer across text distributions, we apply $\eta=2.15$ and $\eta_+=3.30$ without recalibration to $1{,}000$ records from the OpenGen dataset, keeping all other configurations unchanged. The empirical FPRs remain close to the $1\%$ target and the TPRs remain close to those on C4, supporting threshold transfer under the same configuration. BA and EMR are lower than on C4, reflecting the distribution shift between the two corpora under the same fixed detection thresholds.

\textbf{When recalibration is needed.} We identify two cases where the thresholds should be recalibrated. First, in low-entropy generation such as code or mathematical proofs, a small set of tokens and constructions dominates the text, producing a different shard-occupancy distribution under the null hypothesis than general text. Second, the scale of the dispersion score depends on the number of observed tokens. A threshold calibrated on longer texts may therefore be overly conservative for shorter texts, potentially reducing both FPR and TPR. In practice, we recommend calibrating the thresholds separately for each target domain and token budget.

\section{Ablation under a Fixed Green-List Ratio}
\label{app:fixed_gamma}

We further investigate whether the performance improvements of our methods primarily arise from their larger green-list ratios. To remove this factor, we conduct an additional experiment using randomly generated balanced messages with \(b=64\), such that the numbers of \(0\)s and \(1\)s are equal. For MajorMark+, we set \(r=1\), treating the entire message as a single block. Under this setting, RSBH, MajorMark, and MajorMark+ all use the same green-list ratio of \(\gamma=0.5\). We evaluate 100 randomly generated messages, and the results are summarized in Table~\ref{tab:fixed_gamma}.

\begin{wraptable}{r}{0.4\textwidth}
\centering
\caption{Performance under a fixed green-list ratio of \(\gamma=0.5\) with message length \(b=64\). Results are averaged over 100 balanced messages.}
\label{tab:fixed_gamma}
\scalebox{0.8}{
\begin{tabular}{lcc}
\toprule
\textbf{Method} & \textbf {BA} \(\uparrow\) & \textbf{PPL} \(\downarrow\) \\
\midrule
RSBH & \(82.58 \) & \(4.72 \) \\
\rowcolor[rgb]{  .9, .9, .9}MajorMark & \(87.59 \) & \(4.81 \) \\
\rowcolor[rgb]{  .9, .9, .9}MajorMark+ (\(r=1\)) & \(\mathbf{90.67 }\) & \(\mathbf{4.68 }\) \\
\bottomrule
\end{tabular}
}
\end{wraptable}

As shown in Table~\ref{tab:fixed_gamma}, the three methods achieve comparable PPL when the green-list ratio is fixed at \(\gamma=0.5\). In contrast, MajorMark improves BA over RSBH by \(5.01\) percentage points, while MajorMark+ further increases the improvement to \(8.09\) percentage points. These results show that the performance gains of our methods do not arise solely from using a larger green list. The proposed majority bit-aware encoding and the corresponding decoding strategies can provide improved decoding accuracy even when the green-list ratio is controlled.

\section{Pareto Curves of \texorpdfstring{MajorMark$^+$}{MajorMark+}}
\label{apdx: pareto_curves}

\setlength{\intextsep}{0mm}%
\begin{wrapfigure}{r}{0.4\textwidth}
    \centering
    \includegraphics[width=0.7\linewidth]{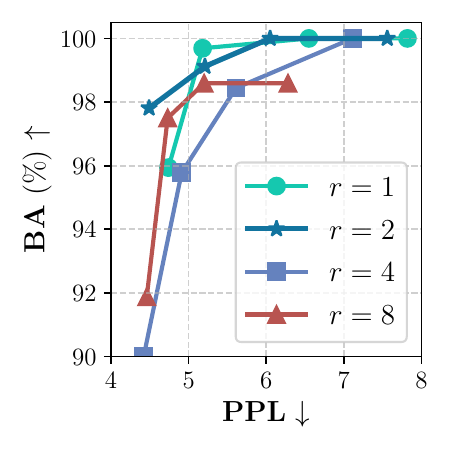}
    \caption{
    Pareto comparison under different $(r,\delta)$ combinations.
    }
    \label{fig:pareto}
\end{wrapfigure}
We plot the Pareto curves of different $(r,\delta)$ combinations in Figure~\ref{fig:pareto}, where the points on each curve from left to right correspond to $\delta =2, 3, 4, 6$. Since lower PPL and higher BA are preferred, curves closer to the upper-left region indicate a better trade-off between text quality and decoding accuracy. Among all evaluated settings, $r=2$ generally lies closer to the Pareto front. In particular, it achieves high BA with relatively low PPL when $\delta$ is small, and reaches nearly perfect BA as $\delta$ increases, while avoiding the larger PPL increase observed in some other settings. This suggests that $r=2$ provides a better balance between watermarked text quality and message decoding accuracy.

\section{Algorithms of MajorMark} \label{sec: algorithm_of_majormark}

\textbf{Encoding.}
We present the detailed encoding procedure of MajorMark in Algorithm~\ref{alg: majormark_encode}. The hash function $\mathtt{Hash}(\cdot)$ can be implemented flexibly; in our implementation, we use the formula $(k \times x_{t-1} \times x_{t-2} + \lambda \times 31) \bmod 2^{64}$, where $k = 15{,}485{,}863$. For the permutation step $\mathtt{Permute(\cdot)}$, we adopt PyTorch’s built-in \texttt{torch.randperm} to generate a permutation of the vocabulary. The $\mathtt{Partition(\cdot)}$ function then divides the permuted vocabulary $\mathcal{V}^\prime$ into $b$ equal-sized shards in order. 

\begin{algorithm}[t]
\caption{The Encoding Function of MajorMark}
\label{alg: majormark_encode}
\SetKwInOut{Input}{Input}\SetKwInOut{Output}{Output}
\Input{User prompt $\mathbf{x}_p$, Maximum length $T$, Message $\mathbf{m} \in \{0,1\}^b$, Secret key $k$, LLM $f$, Vocabulary $\mathcal{V}$, Bias strength $\delta$}
\Output{Watermarked text $\mathbf{x}_g^\prime$}
$\mathbf{x}_{:-1} \leftarrow \mathbf{x}_p$ \\
$\lambda \leftarrow \mathtt{get\_majority\_bit}(\mathbf{m})$ \\
\For{$t = 0$ \KwTo $T-1$}{
    $x_{t-1}, x_{t-2} \leftarrow \mathbf{x}_{t-1}, \mathbf{x}_{t-2}$ \\
    $s \leftarrow \mathtt{Hash}(k, x_{t-1}, x_{t-2}, \lambda)$ \\
    $\mathcal{V}^\prime \leftarrow \mathtt{permute}(\mathcal{V}, s)$ \\
    $\mathrm{shard}_1, \ldots, \mathrm{shard}_b \leftarrow \mathtt{partition}(\mathcal{V}^\prime, b)$ \\
    $\mathcal{G} \leftarrow \emptyset$ \\
    \For{$i = 1$ \KwTo $b$}{
        \If{$m_i == \lambda$}{
            Append $\mathrm{shard}_i$ to $\mathcal{G}$
        }
    }
    $\ell^t \leftarrow f(\mathbf{x}_{:t})$ \tcp*{Get next-token logits}
    \For{$j = 0$ \KwTo $|\mathcal{V}| - 1$}{
        \If{$j \in \mathcal{G}$}{
            $\ell^t_j \leftarrow \ell^t_j + \delta$
        }
    }
    Sample $x_t \sim \mathtt{Softmax}(\ell^t)$ \\
    Append $x_t$ to $\mathbf{x}$
}
\textbf{Return} $\mathbf{x}_{0:T-1}$
\end{algorithm}

\textbf{Decoding.}
The decoding procedure of MajorMark is detailed in Algorithm~\ref{alg: majormark_decode}. To ensure consistency, the hash function $\mathtt{Hash}(\cdot)$ used during decoding must exactly match the one used in encoding. The function $\mathtt{std}(\cdot)$ computes the standard deviation of a numeric array, while $\mathtt{argmax}(\cdot)$ returns the index of the maximum value in a list. We employ the $\mathtt{KMeans(\cdot)}$ clustering algorithm from the scikit-learn library~\citep{scikit-learn}, with the number of clusters set to $2$ and all other parameters set to their default values. The returned cluster $\mathcal{C}$ is the one with the higher mean shard occurrence count. The final message is reconstructed by assigning the majority bit $\lambda$ to shards in $\mathcal{C}$ and the complement bit $1{-}\lambda$ to the remaining shards.

\setlength{\textfloatsep}{10pt}
\begin{algorithm}[t]
\caption{The Decoding Function of MajorMark}
\label{alg: majormark_decode}
\SetKwInOut{Input}{Input}
\SetKwInOut{Output}{Output}
\Input{Secret key $k$, LLM $f$, Vocabulary $\mathcal{V}$, Unverified text $\mathbf{x}$ of length $T$}
\Output{Decoded message $\mathbf{m}^\prime$}
Initialize $\mathtt{occ} \leftarrow$ zero matrix of shape $2 \times b$ \tcp*{Shard-wise token counts for both $\lambda^\prime$}

\For{$\lambda^\prime \in \{0,1\}$}{
    \For{$t=3$ \KwTo $T$}{
        $x_{t-1}, x_{t-2} \leftarrow \mathbf{x}_{t{-}1}, \mathbf{x}_{t{-}2}$ \\
        $s \leftarrow \mathtt{Hash}(k, x_{t-1}, x_{t-2}, \lambda^\prime)$ \\
        $\mathcal{V}^\prime \leftarrow \mathtt{permute}(\mathcal{V}, s)$ \\
        $\mathrm{shard}_1, \dots, \mathrm{shard}_b \leftarrow \mathtt{partition}(\mathcal{V}^\prime, b)$ \\
        \For{$i = 1$ \KwTo $b$}{
            \If{$x_t \in \mathrm{shard}_i$}{
                $\mathtt{occ}[\lambda^\prime][i] \leftarrow \mathtt{occ}[\lambda^\prime][i] + 1$ \\
                \textbf{break}
            }
        }
    }
    $\sigma_{\lambda^\prime} \leftarrow \mathtt{std}(\mathtt{occ}[\lambda^\prime])$
}


$\lambda \leftarrow \mathtt{argmax}(\sigma_0, \sigma_1)$ 

Cluster $\mathtt{occ}[\lambda]$ into 2 groups via $\mathtt{KMeans}(\cdot, 2)$ \\
Let $\mathcal{C}$ be the cluster with higher average count

\For{$j = 1$ \KwTo $b$}{
    \uIf{$j \in \mathcal{C}$}{
        $m_j \leftarrow \lambda$
    }
    \Else{
        $m_j \leftarrow 1 - \lambda$
    }
}

\textbf{Return} $\mathbf{m}^\prime = (m_1, m_2, \dots, m_b)$
\end{algorithm}

\section{Algorithms of \texorpdfstring{MajorMark$^+$}{MajorMark+}}\label{sec: algorithm_of_majormarkplus}
\textbf{Encoding.}
The encoding procedure of MajorMark$^+$ is detailed in Algorithm~\ref{alg: majormarkplus_encode}. We begin by dividing the full message $\mathbf{m} \in {0,1}^b$ into $r$ disjoint blocks using the function $\mathtt{divide}(\cdot)$. For each block $\mathbf{m}_w$, the function $\mathtt{get\_majority\_bit}(\cdot)$ returns both the majority bit $\lambda_w$ and its frequency $h_{\lambda_w}$. The hash function $\mathtt{Hash}(\cdot)$ can be implemented flexibly; in our implementation, we use the formula $(k \times x_{t-1} \times x_{t-2} + \lambda_p \times 31 + h_{\lambda_p} \times 97) \bmod 2^{64}$, where $k = 15{,}485{,}863$. The permutation and partition functions, denoted $\mathtt{Permute(\cdot)}$ and $\mathtt{Partition(\cdot)}$, respectively, are consistent with those used in the original MajorMark encoding.

\setlength{\textfloatsep}{10pt}
\begin{algorithm}[t]
\small
\caption{The Encoding Function of MajorMark$^+$}
\label{alg: majormarkplus_encode}
\SetKwInOut{Input}{Input}
\SetKwInOut{Output}{Output}

\Input{User prompt $\mathbf{x}_p$, maximum generation length $T$, message $\mathbf{m} \in \{0,1\}^b$, secret key $k$, LLM $f$, vocabulary $\mathcal{V}$, bias strength $\delta$, number of blocks $r$}
\Output{Watermarked text $\mathbf{x}_g^\prime$}

$d \leftarrow b/r$ \\

Initialize $\mathbf{x}_g^\prime \leftarrow []$ \\
Initialize $\mathbf{x} \leftarrow \mathbf{x}_p$ \tcp*{Current context}

Divide $\mathbf{m}$ into $r$ equal-sized blocks: 
$\mathbf{m}_1, \dots, \mathbf{m}_r \leftarrow \mathtt{divide}(\mathbf{m}, r)$ \\

\For{$p = 1$ \KwTo $r$}{
    $\lambda_p, h_{\lambda_p} \leftarrow \mathtt{get\_majority\_bit}(\mathbf{m}_p)$
}

\For{$t = 1$ \KwTo $T$}{
        $x_{t-1}, x_{t-2} \leftarrow \mathbf{x}_{t{-}1}, \mathbf{x}_{t{-}2}$ \\
    
    $p \leftarrow ((x_{t-1} + x_{t-2}) \bmod r) + 1$ \tcp*{Select block index}
    
    $s \leftarrow \mathtt{Hash}(k, x_{t-1}, x_{t-2}, \lambda_p, h_{\lambda_p})$ \\
    $\mathcal{V}^\prime \leftarrow \mathtt{permute}(\mathcal{V}, s)$ \\
    $\mathcal{V}_1, \dots, \mathcal{V}_d \leftarrow \mathtt{partition}(\mathcal{V}^\prime, d)$ \\
    
    $\mathcal{G} \leftarrow \emptyset$ \\
    \For{$i = 1$ \KwTo $d$}{
        \If{$(\mathbf{m}_p)_i = \lambda_p$}{
            $\mathcal{G} \leftarrow \mathcal{G} \cup \mathcal{V}_i$
        }
    }
    
    $\ell^t \leftarrow f(\mathbf{x})$ \tcp*{Get next-token logits}
    
    \For{$j = 0$ \KwTo $|\mathcal{V}| - 1$}{
        \If{$j \in \mathcal{G}$}{
            $\ell^t_j \leftarrow \ell^t_j + \delta$
        }
    }
    
    Sample $x_t^\prime \sim \mathtt{Softmax}(\ell^t)$ \\
    Append $x_t^\prime$ to $\mathbf{x}_g^\prime$ \\
    Append $x_t^\prime$ to $\mathbf{x}$
}

\textbf{Return} $\mathbf{x}_g^\prime$
\end{algorithm}

\setlength{\textfloatsep}{10pt}
\begin{algorithm}[t]
\small
\caption{The Decoding Function of MajorMark$^+$}
\label{alg: majormarkplus_decode}
\SetKwInOut{Input}{Input}
\SetKwInOut{Output}{Output}
\Input{Secret key $k$, LLM $f$, Vocabulary $\mathcal{V}$, Unverified text $\mathbf{x}_g^\prime$ of length $T$, Number of blocks $r$}
\Output{Decoded message $\mathbf{m}^\prime$}

Initialize $\mathtt{occ} \leftarrow$ zero matrix of shape $r \times 2 \times (b/r)$ \\
Initialize $\mathtt{occ}[p][\lambda^\prime][h^\prime][i] \leftarrow 0$ \\
$\mathcal{H}_0 \leftarrow \{1, \dots, b/r/2 - 1\}$\\
$\mathcal{H}_1 \leftarrow \{1, \dots, b/r/2\}$\\
\For{$\lambda^\prime \in \{0, 1\}$}{
    \For{$h^\prime \in \mathcal{H}_{\lambda^\prime}$}{
        \For{$t = 3$ \KwTo $T$}{
            $x_{t-1}, x_{t-2} \leftarrow \mathbf{x}_{ t{-}1}, \mathbf{x}_{t{-}2}$ \\
            $s \leftarrow \mathtt{Hash}(k, x_{t-1}, x_{t-2}, \lambda^\prime, h^\prime)$ \\
            $\mathcal{V}^\prime \leftarrow \mathtt{permute}(\mathcal{V}, s)$ \\
            $\mathrm{shard}_1, \dots, \mathrm{shard}_{b/r} \leftarrow \mathtt{partition}(\mathcal{V}^\prime, b/r)$ \\
            $p \leftarrow (x_{t-1} + x_{t-2}) \bmod r$ \\
            \For{$i = 1$ \KwTo $b/r$}{
                \If{$x_t \in \mathrm{shard}_i$}{
                    $\mathtt{occ}[p][\lambda^\prime][i] \leftarrow \mathtt{occ}[p][\lambda^\prime][i] + 1$ \\
                    \textbf{break}
                }
            }
        }
    }
}

\For{$p = 0$ \KwTo $r-1$}{
    Initialize $\texttt{best\_std} \leftarrow -1$, $(\lambda_p, h_{\lambda_p}) \leftarrow (0, 0)$ \\
    \For{$\lambda^\prime \in \{0, 1\}$}{
        \For{$h^\prime \in \mathcal{H}_{\lambda^\prime}$}{
            $\sigma \leftarrow \mathtt{std}(\mathtt{occ}[p][\lambda^\prime][0:b/r])$ \\
            \If{$\sigma > \texttt{best\_std}$}{
                $\texttt{best\_std} \leftarrow \sigma$ \\
                $(\lambda_p, h_{\lambda_p}) \leftarrow (\lambda^\prime, h^\prime)$
            }
        }
    }
}

Initialize $\mathbf{m}^\prime \leftarrow []$ \\
\For{$p = 0$ \KwTo $r-1$}{
    $c \leftarrow \mathtt{occ}[p][\lambda_p]$ \\
    $\mathcal{T}_p \leftarrow \mathtt{Topk}(c, h_{\lambda_p})$  \\
    \For{$j = 0$ \KwTo $b/r - 1$}{
        \If{$j \in \mathcal{T}_p$}{
            $m_{p,j} \leftarrow \lambda_p$
        }
        \Else{
            $m_{p,j} \leftarrow 1 - \lambda_p$
        }
    }
    Append $(m_{p,0}, \dots, m_{p,b/r-1})$ to $\mathbf{m}^\prime$
}

\textbf{Return} $\mathbf{m}^\prime$
\end{algorithm}

\textbf{Decoding.}
The decoding procedure of MajorMark$^+$ is outlined in Algorithm~\ref{alg: majormarkplus_decode}. Since the message $\mathbf{m}$ is partitioned into $r$ blocks, we decode the message block by block. For each block, we consider both possible values of the majority bit hypothesis $\lambda^\prime \in {0, 1}$, along with all feasible values for its frequency $h_{\lambda^\prime}$. Specifically, when $\lambda^\prime = 0$, there are $b/r/2 - 1$ valid values for $h_{\lambda^\prime}$, and when $\lambda^\prime = 1$, there are $b/r/2$ possible values. We store token occurrence counts in a 3D matrix $\mathtt{occ}$ of shape $r \times 2 \times b/r$.

The hash function $\mathtt{Hash}(\cdot)$ used during decoding must exactly match the one used in the encoding process of MajorMark$^+$. For each combination of $(\lambda^\prime, h_{\lambda^\prime})$, we compute the standard deviation of shard-wise token occurrences. The configuration that yields the highest standard deviation is selected, and we recover the corresponding majority bit $\lambda_p$ and frequency $h_{\lambda_p}$ for each block $p \in [r]$. To reconstruct the message, we identify the top-$h_{\lambda_p}$ shards (using $\mathtt{Topk}$) with the highest counts in each block. The majority bit $\lambda_p$ is assigned to these shards, and the remaining shards are assigned the complement bit $1{-}\lambda_p$. Finally, the decoded message $\mathbf{m}^\prime$ is obtained by concatenating all block-wise results.

\section{Theoretical Justification of Clustering-based Decoding} \label{apdx: theo_cluster}

\textbf{Theoretical Explanation of Clustering-Based Decoding.}
Under the ROM assumption, shard selections can be viewed as random draws induced by the watermarked generation distribution. By the Central Limit Theorem~\citep{feller1991introduction}, the observed occurrence count $freq_i$ of each shard $i$ can be approximated by a Gaussian distribution when the number of decoding tokens is sufficiently large. Under majority bit-aware encoding, for a fixed majority bit $\lambda$, shards associated with $\lambda$ receive a larger expected occurrence probability than those associated with $1-\lambda$, provided that the watermark strength satisfies $\delta>0$. Let $\Delta(\delta)>0$ denote the resulting expected per-shard probability gap between the two groups. Let $\mathcal{S}_{\lambda}$ and $\mathcal{S}_{1-\lambda}$ denote the shard groups associated with $\lambda$ and $1-\lambda$, respectively. We define the average occurrence counts of the two groups over the $T$ decoding tokens as
\[
\bar C_{\lambda}
=
\frac{1}{|\mathcal{S}_{\lambda}|}
\sum_{i\in \mathcal{S}_{\lambda}} freq_i,
\qquad
\bar C_{1-\lambda}
=
\frac{1}{|\mathcal{S}_{1-\lambda}|}
\sum_{i\in \mathcal{S}_{1-\lambda}} freq_i,
\]
and define their empirical separation as $D_T=\bar C_{\lambda}-\bar C_{1-\lambda}.
$
Under the induced watermarking distribution, the expected separation grows linearly with the number of decoding tokens: $\mathbb{E}[D_T]=T\cdot \Delta(\delta),
$
while its variance grows at most linearly with $T$:
$\mathrm{Var}(D_T)=O(T).
$
Therefore, the signal-to-noise ratio of the group separation satisfies
$\frac{\mathbb{E}[D_T]}{\sqrt{\mathrm{Var}(D_T)}}=\Omega(\sqrt{T}),$
which increases with the number of decoding tokens. This indicates that, as $T$ becomes sufficiently large, the empirical occurrence statistics of the two shard groups become increasingly separable. Under standard concentration arguments, the probability of group-separation failure, characterized by $D_T\leq 0$, decreases toward zero. This provides an asymptotic consistency explanation for MajorMark: with sufficiently many decoding tokens, the correct shard partition can be recovered with arbitrarily high probability.

\textbf{Error analysis via distribution overlap.}
The above consistency relies on sufficient empirical separation between the two shard groups. In finite-length generations, however, decoding errors may still occur when the empirical distributions of green and red shard counts overlap. This overlap is mainly caused by two factors. \ding{172} \textit{Insufficient text length.} When $T$ is small, the variance of the empirical occurrence counts is large relative to the mean gap between the two shard groups. As a result, green and red shards may not be clearly separated, leading to higher misclassification rates and lower bit accuracy. This explains the performance drop for short outputs, as shown in Figure~\ref{fig:AvaTokens}. \ding{173} \textit{Weak watermark bias.} When $\delta$ is small, the probability gap $\Delta(\delta)$ between green and red shards becomes small. In this case, even though the separation can still improve with a larger $T$, more decoding tokens are required to achieve reliable recovery. With limited text length, the two groups may become nearly indistinguishable, causing $\mathtt{KMeans}$ to behave close to random partitioning. This matches our results in Table~\ref{tab:main_result_table}, where a smaller $\delta$ yields lower decoding accuracy. These analyses explain both the robustness and the finite-sample limitations of the clustering-based decoding design. We leave exact theoretical guarantees for MajorMark$^+$ as an important direction for future work.

\section{Analysis of Robustness under Text Editing Attacks} \label{apdx: theo_robustness_para}

This section analyzes why MajorMark and MajorMark$^+$ can retain part of their watermark signal under standard text-editing attacks and paraphrasing. The stability of the two methods follows from (1) the statistical nature of the noise created by paraphrase and (2) the decoding procedures that remain effective as long as the underlying shard distributions preserve a detectable gap.

\textbf{Statistical behavior of paraphrase noise.} A paraphraser replaces a subset of watermarked tokens with contextually similar alternatives. These replacements produce two effects on the shard-count distribution. \textit{1. Surviving tokens.} paraphrase does not rewrite all positions. Tokens that remain unchanged continue to vote for their correct shards, preserving part of the watermark signal. \textit{2. Replaced tokens.} From the decoder’s viewpoint, replaced tokens behave like random samples from the vocabulary. Due to the majority-bit encoding, the green-shard ratio is at least $0.5$. This ensures that replacements dilute the signal rather than destroying it. In imbalanced cases where $h_\lambda > b/2$, the replacements even introduce a slight bias toward the green component. \textit{3. Overall effect.} The paraphraser weakens the signal but does not erase it. The shard-count distribution remains biased toward the majority-bit shards, maintaining a non-zero gap that decoding algorithms can recover.


\textbf{Robustness of clustering-based decoding in MajorMark.}
Recall that in Appendix~\ref{apdx: theo_cluster}, the occurrence count $freq_i$ of each shard is approximated by a Gaussian distribution under the ROM assumption. The watermark bias $\delta$ increases the expected occurrence probability of green shards over red shards, creating two shard groups with different expected counts. Let $\mathcal{S}_{\lambda}$ and $\mathcal{S}_{1-\lambda}$ denote the shard groups associated with the majority bit $\lambda$ and the minority bit $1-\lambda$, respectively. Their empirical separation can be measured by
$D_T =
\frac{1}{|\mathcal{S}_{\lambda}|}\sum_{i\in \mathcal{S}_{\lambda}} freq_i
-
\frac{1}{|\mathcal{S}_{1-\lambda}|}\sum_{i\in \mathcal{S}_{1-\lambda}} freq_i .$
Under the watermarked generation distribution, $\mathbb{E}[D_T]$ grows linearly with the number of decoding tokens $T$, while its variance grows at most linearly with $T$. Thus, the signal-to-noise ratio of the group separation increases with $T$. Paraphrasing may reduce the effective probability gap between green and red shards, thereby decreasing the empirical separation $D_T$. However, as long as the reduced separation remains larger than the noise scale, the two groups of shard counts remain distinguishable. Since $\mathtt{KMeans}$ only relies on the relative separation between the two empirical groups rather than their exact mean values, it can still recover the correct shard partition when the bimodal structure is preserved. This explains why MajorMark maintains stable decoding performance under paraphrasing.

\textbf{Robustness of block-wise decoding in MajorMark$^+$.}
MajorMark$^+$ extends MajorMark with a block-wise structure. Paraphrasing can modify the context used to compute the block index $i$, causing tokens to be assigned to the wrong blocks during decoding. This does not form a structural weakness for the following reasons. \ding{172} \textit{Uniform noise across blocks.} When context tokens are altered, the resulting block index becomes random. The contribution of the replaced token is therefore spread uniformly across all blocks rather than targeted at a specific block. \ding{173} \textit{Graceful degradation.} A wrongly assigned token does not vote against the correct block. It simply becomes noise. Because the green-shard ratio is high, the correctly synchronized tokens provide enough evidence for recovering the block message. \ding{174} \textit{Independence across blocks.} Each block is decoded independently. Noise affecting one block does not propagate to the others. This prevents local errors from causing global decoding failure.

\textbf{Summary.}
Under mild paraphrasing, MajorMark and MajorMark$^+$ can retain part of the watermark signal as long as a detectable statistical gap remains between the shard distributions. This analysis does not extend to adaptive paraphrasers specifically optimized to remove watermark signals, as evaluated below.

\paragraph{Adaptive Watermark-Removal Attack.}
The above analysis focuses on standard text-editing attacks and mild paraphrasing. To further evaluate robustness against a stronger adaptive attack, we test the watermark-removal paraphraser proposed by~\cite{diaa2025optimizing}, WM-Removal-KGW-Llama-2-7B, which is specifically tuned against KGW-family watermarks. We evaluate all methods on LLaMA2-7B, and report the results in Table~\ref{tab:adaptive_attack}.

\begin{wraptable}{r}{0.4\textwidth}
\centering
\caption{Robustness under the adaptive watermark-removal paraphraser WM-Removal-KGW-Llama-2-7B. The $\pm$ values denote 95\% CIs.}
\label{tab:adaptive_attack}
\scalebox{0.8}{
\begin{tabular}{lcc}
\toprule
\textbf{Method} & \textbf{BA} \(\uparrow\) & \textbf{PPL} \(\downarrow\) \\
\midrule
RSBH & \(53.22_{\pm 1.98}\) & \(4.70 _{\pm 0.15}\) \\
MPAC & \(\mathbf{59.88 }_{\pm 1.81}\) & \(4.86 _{\pm 0.14}\) \\
\rowcolor[rgb]{  .9, .9, .9}MajorMark & \(58.28 _{\pm 1.85}\) & \(4.60 _{\pm 0.15}\) \\
\rowcolor[rgb]{  .9, .9, .9}MajorMark$^+$ & \(55.53 _{\pm 1.99}\) & \(\mathbf{4.52 }_{\pm 0.12}\) \\
\bottomrule
\end{tabular}
}
\end{wraptable}

Under this tailored attack, the BA of all methods drops to \(53.22\%\)--\(59.88\%\), approaching random guessing. This result shows that the robustness observed under standard paraphrasing does not extend to an adaptive paraphraser specifically optimized to remove KGW-style watermark signals. The limitation is not specific to majority bit-aware encoding, as all tested KGW-family methods experience substantial degradation. Improving robustness against such adaptive watermark-removal attacks remains an important direction for future work.

\section{Failure Analysis of Majority-Bit Recovery}
\label{app:lambda_failure}

MajorMark first recovers the majority bit \(\lambda\) by comparing the dispersion of the shard-wise token occurrence counts under the two candidate values \(\lambda \in \{0,1\}\). Since \(\lambda\) is also an input to the hash function, an incorrect recovery of \(\lambda\) produces token-to-shard mappings that differ from those used during encoding. Consequently, a single error in \(\lambda\) can cause correlated errors across multiple decoded bits. To evaluate this failure mode, we conduct a controlled experiment under increasingly weak watermark signals by reducing the watermarking bias from the default \(\delta=2.0\) to \(1.5\) and \(1.0\). We evaluate 100 randomly generated messages and report PPL, BA, the majority-bit failure rate (FR), defined as the fraction of messages for which \(\lambda\) is incorrectly recovered, and wrong-case bit accuracy (WBA), defined as the average BA among messages with incorrect \(\lambda\).

\begin{wraptable}{r}{0.5\textwidth}
\centering
\caption{Failure analysis of majority-bit recovery in MajorMark under weak watermarking biases. The $\pm$ values denote 95\% CIs.}
\label{tab:lambda_failure}
\resizebox{0.9\linewidth}{!}{
\begin{tabular}{c|c|c|c|c}
\toprule
\(\delta\) & \textbf{PPL} \(\downarrow\) & \textbf{BA} \(\uparrow\) & \textbf{FR} & \textbf{WBA} \\
\midrule
2.0 & \(5.48 _{\pm 0.17}\) & \(97.06 _{\pm 0.84}\) & 0/100 & -- \\
1.5 & \(4.34 _{\pm 0.12}\) & \(87.84 _{\pm 1.66}\) & 0/100 & -- \\
1.0 & \(4.06 _{\pm 0.11}\) & \(81.19 _{\pm 2.12}\) & 1/100 & 37.50 \\
\bottomrule
\end{tabular}
}
\end{wraptable}

As shown in Table~\ref{tab:lambda_failure}, we observe no majority-bit recovery failures at \(\delta \geq 1.5\). At \(\delta=1.0\), one failure occurs among 100 messages. In this failed case, BA drops to \(37.50\%\), confirming that an incorrect \(\lambda\) can introduce correlated, message-wide decoding errors. However, the dispersion difference between the two candidate values of \(\lambda\) is only \(0.19\), which is substantially below the calibrated detection threshold. Therefore, this case would be rejected as an unreliable watermark rather than accepted as a confident attribution. A checksum or error-correcting code (ECC) can provide an additional safeguard against this failure mode. For example, the decoder could retain both candidate values of \(\lambda\) and use checksum or codeword validity to select between them, while ECC could further correct residual bit errors. We leave the joint design of majority-bit recovery and error-correcting codes to future work.

\section{Impact of Message Imbalance on Decoding Accuracy}
\label{apdx: bit_imbalance}

This section analyzes how the distribution of bit values in the embedded message affects the decoding behavior of MajorMark and MajorMark$^+$. In both methods, the embedded message determines the construction of green and red shards. When the message is highly imbalanced, i.e., one bit value appears much more frequently than the other, the number of green shards can increase and the green list ratio $\gamma$ approaches $1$. Under a fixed text length $T$ and a fixed watermarking bias $\delta$, this can weaken the watermark signal. Due to the shift invariance of the softmax, adding the same bias to almost all tokens leads to limited relative change in the output distribution. Meanwhile, the probability mass gained by green tokens is distributed across many green shards, reducing the expected count per green shard and weakening the separation between green and red shards.

However, this scenario also provides a compensatory advantage. When $\gamma$ becomes large, the red list becomes a small fraction of the vocabulary. In this case, increasing $\delta$ mainly penalizes a small subset of red tokens and therefore causes less degradation to text quality. As a result, a larger $\delta$ can be used to restore the separability between green and red shard distributions without severely increasing PPL. This effect applies to both MajorMark and MajorMark$^+$. In the following experiment, we use MajorMark$^+$ as a representative case to empirically demonstrate this behavior.

\setlength{\intextsep}{0mm}%
\begin{wraptable}{r}{0.4\textwidth}
    \centering
    \caption{
    Impact of watermarking bias $\delta$ under different message imbalance rates.
    }
    \label{tab:imbalance_delta}
    \resizebox{0.9\linewidth}{!}{
    \begin{tabular}{c|cc}
        \toprule
        \textbf{$\delta$} 
        & \textbf{$h_\lambda/b=71.88\%$} 
        & \textbf{$h_\lambda/b=81.25\%$} \\
        \midrule
        $2$ & $(4.06, 85.62)$ & $(4.09, 80.16)$ \\
        $4$ & $(4.89, 97.66)$ & $(4.39, 96.88)$ \\
        $6$ & $(5.90, 100.00)$ & $(4.91, 100.00)$ \\
        $8$ & $(6.71, 100.00)$ & $(5.54, 99.69)$ \\
        \bottomrule
    \end{tabular}
    }
\end{wraptable}
To verify this effect, we conduct experiments with MajorMark$^+$ under message imbalance rates $h_\lambda/b$ of $71.88\%$ and $81.25\%$, corresponding to $23$ and $26$ bits out of a $b=32$ message matching $\lambda$, respectively. Each entry in Table~\ref{tab:imbalance_delta} reports $(\text{perplexity}, \text{bit accuracy})$. As shown in Table~\ref{tab:imbalance_delta}, under a fixed small bias of $\delta=2$, a higher imbalance rate leads to lower BA. Specifically, the BA decreases from $85.62\%$ at $h_\lambda/b=71.88\%$ to $80.16\%$ at $h_\lambda/b=81.25\%$. This confirms that message imbalance weakens the per-shard watermark signal when the bias is small. However, the results also show that a higher imbalance rate provides stronger tolerance to a larger $\delta$. For example, when $\delta=6$, both imbalance settings achieve $100.00\%$ BA, while the $81.25\%$ imbalance setting achieves a lower PPL of $4.91$, compared with $5.90$ under the $71.88\%$ imbalance setting. Similarly, when $\delta=8$, the $81.25\%$ imbalance setting still maintains a relatively low PPL of $5.54$ and a BA of $99.69\%$.

These results show that message imbalance has two effects for both MajorMark and MajorMark$^+$. On the one hand, it weakens the per-shard decoding signal under a small fixed bias. On the other hand, it increases the green list ratio $\gamma$, which allows the use of a larger $\delta$ with less impact on text quality. Therefore, both MajorMark and MajorMark$^+$ can recover strong decoding accuracy under imbalanced messages by increasing $\delta$, while still maintaining reasonable text quality.

\section{Broader Impacts}\label{apdx:boarderimpact}

Our methods improve multi-bit watermarking for LLM-generated text by achieving a better balance between text quality and message decoding accuracy. This can support practical traceability in LLM services, where service providers may need to embed user identifiers, model versions, or other metadata into generated outputs. Such traceability can help with provenance tracking, intellectual property protection, and misuse investigation when generated text is leaked, redistributed, or used in harmful contexts. However, traceable watermarking also introduces potential risks. If embedded messages are directly linked to user identities, they may raise privacy concerns. In practice, watermark messages should be anonymized, encrypted, or otherwise protected. In addition, watermark detection may be affected by editing, paraphrasing, or other perturbations, and therefore should not be treated as the only evidence in high-stakes cases. Responsible deployment should include clear disclosure, privacy-aware message design, and careful interpretation of detection results.

\section{Use of LLM Statement}
We used LLM solely for grammar checking and polishing the writing of this manuscript.

\begin{table*}[ht]
    \centering
    \caption{Generated text under different methods.}
    \label{tab:generated_text}
    \textbf{Prompt}: \textit{\promptText  [...]} \\
    \renewcommand{\arraystretch}{1.2}
    \scalebox{0.7}{
    \begin{tabular}{l p{15cm} c}  
        \toprule
        \textbf{Method} & \textbf{Generated Text} & \textbf{PPL}$\downarrow$  \\
        \midrule
        No Watermark & short film.
Bryan, who is a co-owner of the MLS team Los Angeles Football Club, has said he wants to open a youth soccer academy.
He is also the co-owner of Granite Hill Capital Partners, a venture capital firm.
His investment firm Easy Street has invested in more than 150 startups, according to Forbes.
Sports Academy is planning to open a new location in the Valley in 2020.
It's unclear if the new facility will be rebranded as Mamba Sports Academy.
Sports reporter Nathaniel Percy contributed to this report.
This article is written by Nathaniel Percy from Ventura County Star, Thousand Oakes and was legally licensed via the Tribune Content Agency through the NewsCred publisher network. Please direct all licensing questions to legal@newscred.com.
© Copyright 2019 Ventura County Star. All rights reserved. This material may not be published, broadcast, rewritten or redistributed.
Ventura County Star is the leading media company in Ventura County, California. The & $3.01$  \\ \\

        MPAC & \mpacfour & $8.58$  \\ \\

        RSBH & \rsbhfour & $6.09$  \\ \\

        MajorMark & \mamarkdeltatwo & $5.74$ \\ \\
        
        MajorMark$^+$ & \plusfour & $4.47$ \\
        \bottomrule
    \end{tabular}
    }
\end{table*}

\clearpage

\section{Proof of the Guaranteed Green List Size of MajorMark}
\label{apdx: proof_of_green_list}

\textbf{Assumption.} 
We begin with an assumption on the distribution of bits in the embedded message.

\begin{assumption}[Uniform Bit Distribution]\label{assm: bit_distribution}
Each bit in the message $\mathbf{m} \in \{0,1\}^b$ is sampled independently and uniformly at random. That is, for all $i \in \{1, \dots, b\}$, $\Pr(m_i = 0) = \Pr(m_i = 1) = 0.5, \ \text{and } m_i \perp m_j \text{ for } i \neq j.$
\end{assumption}

\textbf{Preliminaries.}
We will make use of two theoretical tools: the De Moivre--Laplace Theorem for approximating the binomial distribution and a result on the mean absolute deviation of a normal distribution.

\begin{theorem}[De Moivre--Laplace Central Limit Theorem]\label{them: demoivre}
Let \( X_n \sim \mathrm{Binomial}(n, p) \). Then for any real numbers \( z_1 < z_2 \), the following convergence holds:
\[
\lim_{n \to \infty} \Pr\left( z_1 \leq \frac{X_n - np}{\sqrt{np(1-p)}} \leq z_2 \right) = \int_{z_1}^{z_2} \frac{1}{\sqrt{2\pi}} e^{-z^2/2} \, dz.
\]
Equivalently, the standardized variable
$Z_n := \frac{X_n - np}{\sqrt{np(1-p)}}$
converges in distribution to a standard normal variable: $
Z_n \xrightarrow{d} \mathcal{N}(0,1), \text{as } n \to \infty.$
As a consequence, for large \( n \), the binomial distribution can be approximated as: $
X_n \approx \mathcal{N}(np,\, np(1-p)).$
\end{theorem}

\begin{proposition}[Mean Absolute Deviation of Normal Distribution]\label{prop: mad_ofND}
Let \( Z \sim \mathcal{N}(\mu, \sigma^2) \). Then the expected absolute deviation from the mean is given by:
\[
\mathbb{E}[\, |Z - \mu| \,] = \sigma \sqrt{\frac{2}{\pi}}.
\]
\begin{proof}
We evaluate the expectation by integrating:
\[
\mathbb{E}[\, |Z - \mu| \,] = \int_{-\infty}^{\infty} |z - \mu| \cdot \frac{1}{\sqrt{2\pi\sigma^2}} e^{-(z - \mu)^2 / (2\sigma^2)} \, dz.
\]
Applying the change of variable \( x = \frac{z - \mu}{\sigma} \), we obtain:
\[
\mathbb{E}[\, |Z - \mu| \,] = \sigma \int_{-\infty}^{\infty} |x| \cdot \frac{1}{\sqrt{2\pi}} e^{-x^2/2} \, dx = \sigma \sqrt{\frac{2}{\pi}},
\]
which completes the proof.
\end{proof}
\end{proposition}

\textbf{Main Proof.}
We now prove the expected green list size under Assumption~\ref{assm: bit_distribution}.

\begin{proof}
Let $\mathbf{m} = (m_1, m_2, \dots, m_b) \in \{0,1\}^b$ be the embedded message. During watermark encoding, the vocabulary $\mathcal{V}$ is partitioned into $b$ equal-sized shards, each of size $|\mathcal{V}|/b$, with each shard corresponding to a bit $m_i$ in the message. Hereafter, we assume the vocabulary size is perfectly divisible by the message length. Let $h_0$ and $h_1$ denote the number of zeros and ones in $\mathbf{m}$, respectively. Then \( h_0 + h_1 = b \), and under Assumption~\ref{assm: bit_distribution}, we have \( h_1 \sim \mathrm{Binomial}(b, 0.5) \).

Recall that we define the majority bit \( \lambda \in \{0,1\} \) as:
$\lambda = 
\begin{cases}
1 & \text{if } h_1 \ge h_0, \\
0 & \text{otherwise}.
\end{cases}$
The green list \( \mathcal{G} \) is constructed as the union of shards whose corresponding bits equal \( \lambda \). Therefore, the size of the green list is given by: $|\mathcal{G}| = \max(h_0, h_1) \cdot \frac{|\mathcal{V}|}{b}.$
Defining the green list ratio as \( \gamma := \frac{|\mathcal{G}|}{|\mathcal{V}|} \), we observe that:
\[
\gamma = \frac{\max(h_0, h_1)}{b} \ge \frac{1}{2},
\]
since the maximum of two nonnegative numbers summing to $b$ is always at least $b/2$.

We now proceed to compute the expected value of \( \gamma \), or equivalently, the expected size of \( |\mathcal{G}| \). Taking expectation over all messages, we have:
\[
\mathbb{E}_\mathbf{m}[|\mathcal{G}|] = \mathbb{E}\left[\max(h_1, b - h_1)\right] \cdot \frac{|\mathcal{V}|}{b}.
\]

Noting that: $\max(h_1, b - h_1) = \frac{b}{2} + \left|h_1 - \frac{b}{2}\right|,$
we obtain: $\mathbb{E}\left[\max(h_1, b - h_1)\right] = \frac{b}{2} + \mathbb{E}\left[\left|h_1 - \frac{b}{2}\right|\right].$

By Theorem~\ref{them: demoivre}, for large \( b \), the binomial variable \( h_1 \sim \mathrm{Binomial}(b, 0.5) \) is approximated by a normal distribution: $h_1 \approx \mathcal{N}\left(\frac{b}{2}, \frac{b}{4}\right).$
Then, by Proposition~\ref{prop: mad_ofND}, the expected absolute deviation satisfies:
$\mathbb{E}\left[\left|h_1 - \frac{b}{2}\right|\right] \approx \sqrt{\frac{b}{4}} \cdot \sqrt{\frac{2}{\pi}} = \sqrt{\frac{b}{2\pi}}.$

Substituting into the previous expression yields:
\[
\mathbb{E}_\mathbf{m}[|\mathcal{G}|] \approx \left( \frac{b}{2} + \sqrt{\frac{b}{2\pi}} \right) \cdot \frac{|\mathcal{V}|}{b} = \left( \frac{1}{2} + \frac{1}{\sqrt{2\pi b}} \right) |\mathcal{V}|.
\]

Finally, we conclude that the expected green list ratio is:
\[
\mathbb{E}_\mathbf{m}[\gamma] = \frac{\mathbb{E}_\mathbf{m}[|\mathcal{G}|]}{|\mathcal{V}|} \approx \left( \frac{1}{2} + \frac{1}{\sqrt{2\pi b}} \right),
\]
which completes the proof.
\end{proof}

\section{Proof of the Guaranteed Green List Size of \texorpdfstring{MajorMark$^+$}{MajorMark+}}\label{apdx: proof_of_green_list_famarkplus}

\begin{proof}
MajorMark$^+$ divides the message $\mathbf{m} \in \{0,1\}^b$ into $r$ equal-sized blocks $\{ \mathbf{m}_1, \dots, \mathbf{m}_r \}$, where each block $\mathbf{m}_j$ has length $b/r$. For each block, a green list $\mathcal{G}_j$ is independently constructed over the full vocabulary $\mathcal{V}$ using the same \textit{majority bit-aware encoding} rule as in MajorMark.

Let \( \gamma_j := |\mathcal{G}_j| / |\mathcal{V}| \) denote the green list ratio for block \( j \). As shown in the previous proof, for any message block of length \( b/r \), the following lower bound always holds:
\[
\gamma_j \geq \frac{1}{2}, \quad \text{for all } j \in \{1, \dots, r\}.
\]

We define the overall green list ratio $\gamma$ as:
$\gamma := \frac{1}{r} \sum_{j=1}^{r} \gamma_j.$ As a result, we have:
$\gamma \geq \frac{1}{r} \sum_{j=1}^{r} \frac{1}{2} = \frac{1}{2},$
which guarantees that MajorMark$^+$ always yields a green list containing at least half of the vocabulary.

We now proceed to analyze the expected value of \( \gamma \). From Theorem~\ref{them:famark_gls}, we know that for each block of length \( b/r \), the expected green list ratio is:
\[
\mathbb{E}_{\mathbf{m}_j}[\gamma_j] \approx \frac{1}{2} + \frac{1}{\sqrt{2\pi (b/r)}}, \quad \forall j.
\]

Hence, the expected value of the overall green list ratio is:
\begin{align*}
    \mathbb{E}_{\mathbf{m}}[\gamma] = \mathbb{E}_{\mathbf{m}}\left[ \frac{1}{r} \sum_{j=1}^{r} \gamma_j \right] &= \frac{1}{r} \sum_{j=1}^{r} \mathbb{E}_{\mathbf{m}_j}[\gamma_j] \\ &\approx \left( \frac{1}{2} + \frac{1}{\sqrt{2\pi (b/r)}} \right).
\end{align*}
Multiplying by \( |\mathcal{V}| \), we also obtain the expected green list size:
\[
\mathbb{E}_{\mathbf{m}}[|\mathcal{G}|] = \mathbb{E}_{\mathbf{m}}[\gamma] \cdot |\mathcal{V}| \approx \left( \frac{1}{2} + \frac{1}{\sqrt{2\pi (b/r)}} \right) |\mathcal{V}|,\]
which completes the proof.
\end{proof}

\clearpage






\end{document}